\documentclass[11pt]{article}

\usepackage[final]{acl}

\usepackage{times}
\usepackage{latexsym}

\usepackage[T1]{fontenc}

\usepackage[utf8]{inputenc}

\usepackage{microtype}

\usepackage{inconsolata}
\usepackage{graphicx}
\usepackage{amsmath}    
\usepackage{graphicx}   
\usepackage{hyperref}   
\usepackage{cleveref}   
\usepackage[table]{xcolor} 
\usepackage{amsfonts}
\usepackage{bbm}
\usepackage{algorithm}
\usepackage{algorithmic}
\usepackage{enumitem}
\usepackage{multirow}
\usepackage{booktabs}
\usepackage[most]{tcolorbox}
\usepackage{fontawesome}
\usepackage[normalem]{ulem}
\useunder{\uline}{\ul}{}
\usepackage{subfigure}
\usepackage{tabularx}

\useunder{\uline}{\ul}{}
\definecolor{skyblue}{RGB}{0,120,215}

\newcounter{mycounter} 

\usepackage{tcolorbox}
\newcommand{\findingbox}[2][]{
    \refstepcounter{mycounter} 
    \if\relax\detokenize{#1}\relax\else\label{#1}\fi 
    \begin{tcolorbox}[colframe=black,
                      arc=1pt,
                      boxsep=-2pt,
                      before skip=5pt,  
                      after skip=5pt,   
                      ]
        \noindent{\textbf{\textit{Finding \themycounter.}}} #2 
    \end{tcolorbox}
}

\title{Thinking-Based Non-Thinking: Solving the Reward Hacking Problem\\ in Training Hybrid Reasoning Models via Reinforcement Learning}

\author{
Siyuan Gan$^{1}$,
Jiaheng Liu$^{1}$,
Boyan Wang$^{1}$,
Tianpei Yang$^{1}$,
Runqing Miao$^{3}$,
Yuyao Zhang$^{3}$, \\
{\bf Fanyu Meng$^{3}$, Junlan Feng$^{3}$, Linjian Meng$^{\dag 2}$, Jing Huo$^{\dag 1}$, Yang Gao$^{1}$} \\
$^1$State Key Laboratory of Novel Software Technology, Nanjing University, Nanjing, China \\
$^2$Shanghai Artificial Intelligence Laboratory, Shanghai, China\\
$^3$Jiutian Research, Beijing, China\\
\{gansiyuan, menglinjian\}@smail.nju.edu.cn\\
\{liujiaheng, boyanwang, tianpei.yang, huojing, gaoy\}@nju.edu.cn\\
\{miaorunqing, zhangyuyao, mengfanyu, fengjunlan\}@chinamobile.com\\
}
\makeatletter
\def\blfootnote#1{%
  \begin{NoHyper}%
  \xdef\@thefnmark{}%
  \@footnotetext{#1}%
  \end{NoHyper}%
}
\makeatother
\begin{document}
\maketitle

\begin{abstract}
Large reasoning models (LRMs) have attracted much attention due to their exceptional performance. However, their performance mainly stems from thinking, a long Chain of Thought (CoT), which significantly increase computational overhead. 
To address this overthinking problem, existing work focuses on using reinforcement learning (RL) to train hybrid reasoning models that  automatically decide whether to engage in thinking or not based on the complexity of the query. 
Unfortunately, using RL will suffer the the reward hacking problem, e.g., the model engages in thinking but is judged as not doing so, resulting in incorrect rewards.
To mitigate this problem, existing works either employ supervised fine-tuning (SFT), which incurs high computational costs, or enforce uniform token limits on non-thinking responses, which yields limited mitigation of the problem.
In this paper, we propose \textit{Thinking-Based Non-Thinking} (TNT). It does not employ SFT, and sets different maximum token usage for responses not using thinking across various queries by leveraging information from the solution component of the responses using thinking. Experiments on five mathematical benchmarks demonstrate that TNT reduces token usage by around $50\%$ compared to DeepSeek-R1-Distill-Qwen-1.5B/7B and DeepScaleR-1.5B, while significantly improving accuracy. In fact, TNT achieves the optimal trade-off between accuracy and efficiency among all tested methods. Additionally, the probability of reward hacking problem in TNT’s responses, which are classified as not using thinking, remains below $10\%$ across all tested datasets.
\end{abstract}
\blfootnote{$\dag$ Corresponding author.}

\section{Introduction}\label{sec:Introduction}

Large reasoning models (LRMs)~\citep{xu2025towards}, such as DeepSeek-R1~\citep{guo2025deepseek} and OpenAI o1~\citep{jaech2024openai}, have recently become a central research topic due to their superior capabilities. However, their performance gains rely on extended chain-of-thought (CoT)~\citep{wei2022chain}—also referred to as \textit{thinking}. This reliance leads to protracted and repetitive outputs, resulting in a substantial increase in inference overhead and latency~\citep{li2025system,chen2024not}, a phenomenon referred to as the overthinking problem~\citep{sui2025stop,qu2025survey}.

\begin{figure*}[t]
  \centering
  \includegraphics[width=0.8\linewidth]{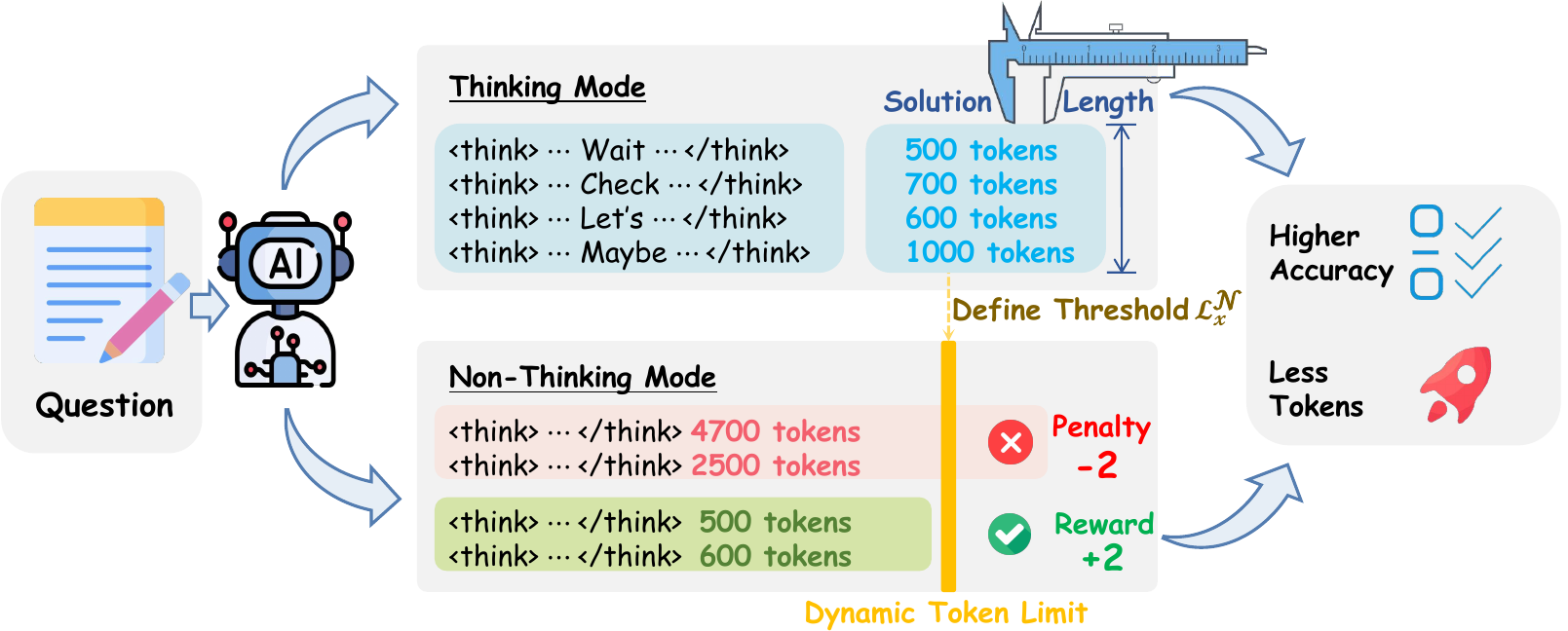}
  \vspace{-0.35cm}
  \caption{Overview of TNT.
  }
  \label{fig:TNT-main}
  \vspace{-0.25cm}
\end{figure*}

\begin{figure*}[t]
 \centering
 \subfigure{
 \begin{minipage}[b]{\linewidth}
  \includegraphics[width=1\linewidth]{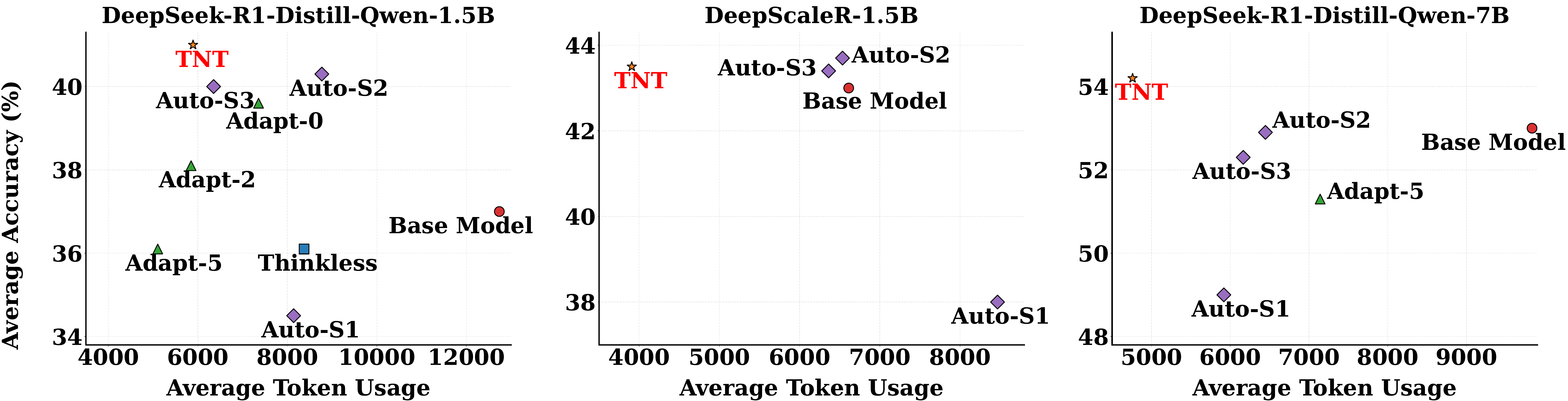}
  \centering 
 \end{minipage}
 }\vspace{-10pt}
 \caption{Average accuracy and token usage comparison across different hybrid reasoning model training methods on mathematical benchmarks. We only presented the evaluation results of their open-source checkpoints while some of these methods lack the trained checkpoints based on DeepScaleR-1.5B, and DeepSeek-R1-Distill-Qwen-7B. Adapt-$x$ refers to AdaptThink with $\delta=x*0.01$. Auto-S$x$ refers to AutoThink-Stage $x$. We also follow these abbreviation conventions in the table below.}
\label{fig:acc_token_sandian}
\vspace{-0.4cm}
\end{figure*}

To tackle the overthinking problem, recent works increasingly explore hybrid reasoning models that can dynamically determine whether to activate the thinking mode~\citep{zhang2025adaptthink,fang2025thinkless,tu2025learning,lou2025adacot,jiang2025_LHRMs,chen2025pangu,zhan2025kat}. The alternative,
where the model directly outputs solution without thinking, is referred to as the non-thinking mode. 
Among these works, one of the most commonly used method involves employing
reinforcement learning (RL), with a higher reward allocated to correct answer of the non-thinking mode than that of the thinking mode~\citep{zhang2025adaptthink,fang2025thinkless,tu2025learning}. Unfortunately, the RL training method suffer from reward hacking problem that reduces the performance. 
Specially, models embed a significant portion of its reasoning process within the non-thinking response~\citep{zhang2025adaptthink,tu2025learning}. By doing so, the model leverages the thinking process to arrive at the correct answer while securing additional rewards for following the non-thinking format. This leads to increased rewards, but results in a significant increase in tokens for non-thinking mode responses.

One approach to mitigate this reward hacking problem is to use supervised fine-tuning (SFT) with a dataset that is significantly larger than the RL dataset, thereby fixing the model's output, as did in Thinkless~\citep{fang2025thinkless}. However, this also 
incurs a substantial increase in computational overhead. To avoid such high computational overhead, AdaptThink~\citep{zhang2025adaptthink} set a smaller maximum token usage for the non-thinking mode than that for the thinking mode. This way, the response judged in the non-thinking mode cannot increase token usage to actually operate in thinking mode and obtain a higher reward. Unfortunately, \citet{zhang2025adaptthink} set the maximum token usage for the non-thinking mode to be uniform across queries, resulting in only limited mitigation of reward hacking problem. 
This is because applying such a uniform maximum may result in poor detection of reward hacking problem because thinking mode tokens for simple queries can be fewer than non-thinking mode tokens for complex ones, making uniform limits ineffective (see details in \Cref{subsec:RL-only training method}).

To set the maximum token usage across different queries, we propose \textit{Thinking-Based Non-Thinking} (TNT). As demonstrated in \Cref{fig:TNT-main}, 
TNT employs the solution component of the thinking mode responses (i.e., the tokens following </think> are highlighted in blue as shown in \Cref{fig:TNT-main}) to set the maximum token usage for the non-thinking mode. Specifically, LRMs' thinking mode has been trained 
to ensure the solution component of the thinking mode responses does not involve thinking, such as DeepSeek-R1. Therefore, this component will not differ significantly from the output of the non-thinking mode as the definition of the latter is that outputting solution without thinking. We can use the token usage of this component to set the maximum token usage for the non-thinking mode.

Consistent with previous hybrid reasoning model training methods~\citep{zhang2025adaptthink,tu2025learning,fang2025thinkless}, we employ DeepSeek-R1-Distill-Qwen-1.5B/7B and DeepScaleR-1.5B as base models to evaluate TNT on five standard mathematical benchmarks: AIME24, AIME25, Minerva, AMC23, and Olympiad. The results show that TNT consistently reduces token usage by approximately $50\%$ relative to base models, while substantially improving accuracy. 
Compared to other hybrid reasoning model training methods, TNT achieves the optimal balance between accuracy and token usage as Figure~\ref{fig:acc_token_sandian}. 
Furthermore, the probability of reward hacking problem in TNT
remains below $10\%$ across all tested datasets.

\section{Preliminary}\label{sec:Preliminary}

Now, we introduce the LRMs. Consider a LRM parameterized by the parameter $\theta$ and denoted by $\pi_\theta$. Given a prompt $x = [x_1, \ldots, x_n, \text{<think>}]$, where $[x_1, \ldots, x_n]$ represents the query and the special token \text{<think>} means the start of thinking. For this prompt, this LRM generates a response $y = [y_1, \ldots, y_{\tau}, \text{</think>}, y_{\tau+2}, \ldots, y_m]$. Here, $[y_1, \ldots, y_{\tau}]$ corresponds to thinking, which is a long CoT consisting of constant exploration, reflection, and self-verification. The token \text{</think>} marks the end of thinking. The remaining sequence, $[y_{\tau+2}, \ldots, y_m]$, denotes the final solution, which only includes the correct steps to solve the problem and the final answer. From the perspective of probability theory, the response $y$ is a sample drawn from the conditional probability distribution $\pi_\theta(\cdot | x)$. Since $y$ is generated in an auto-regressive way, the conditional probability $\pi_\theta(y | x)$ is:
\begin{equation}
\setlength{\abovedisplayskip}{2pt}
\setlength{\belowdisplayskip}{2pt}
\setlength{\abovedisplayshortskip}{2pt}
\setlength{\belowdisplayshortskip}{2pt}
\pi_\theta(y | x) = \prod_{t=1}^m \pi_\theta(y_t | x, y_{<t}).
\end{equation}

In this paper, we follow the approach in \citet{zhang2025adaptthink,tu2025learning,lou2025adacot}\footnote{In \citet{tu2025learning}, they use the \textit{ellipsis prompt} to make $x = [x_1, \ldots, x_n, \text{<think>},\text{\textbackslash  n},\text{...},\text{\textbackslash n}]$, which enables sampling the non-thinking mode without using the off-policy sampling.}, referring to $[y_1, \ldots, y_{\tau}] \neq \emptyset$ as the thinking mode and $[y_1, \ldots, y_{\tau}] = \emptyset$ as the non-thinking mode. We use this approach to distinguish between thinking and non-thinking modes because it does not require modifications to the tokenizer, whereas the methods used in other works do require modifications to the tokenizer~\citep{fang2025thinkless,jiang2025_LHRMs,zhan2025kat}. For example, \citet{fang2025thinkless} determine whether the response is thinking mode or non-thinking mode based on whether the first token of the response $y$ is $\text{<short>}$. Unfortunately, $\text{<short>}$ is usually not in the tokenizer of LRMs, so we need to modify the tokenizer.

\section{Motivation}\label{subsec:Motivation}

We now provide a detailed introduction to the reward hacking problem in training hybrid reasoning models by using RL with a higher reward allocated to the correct answer of the non-thinking mode compared to that of the thinking mode~\citep{zhang2025adaptthink,fang2025thinkless,tu2025learning}.

\begin{figure}[t]
  \includegraphics[width=0.95\columnwidth]{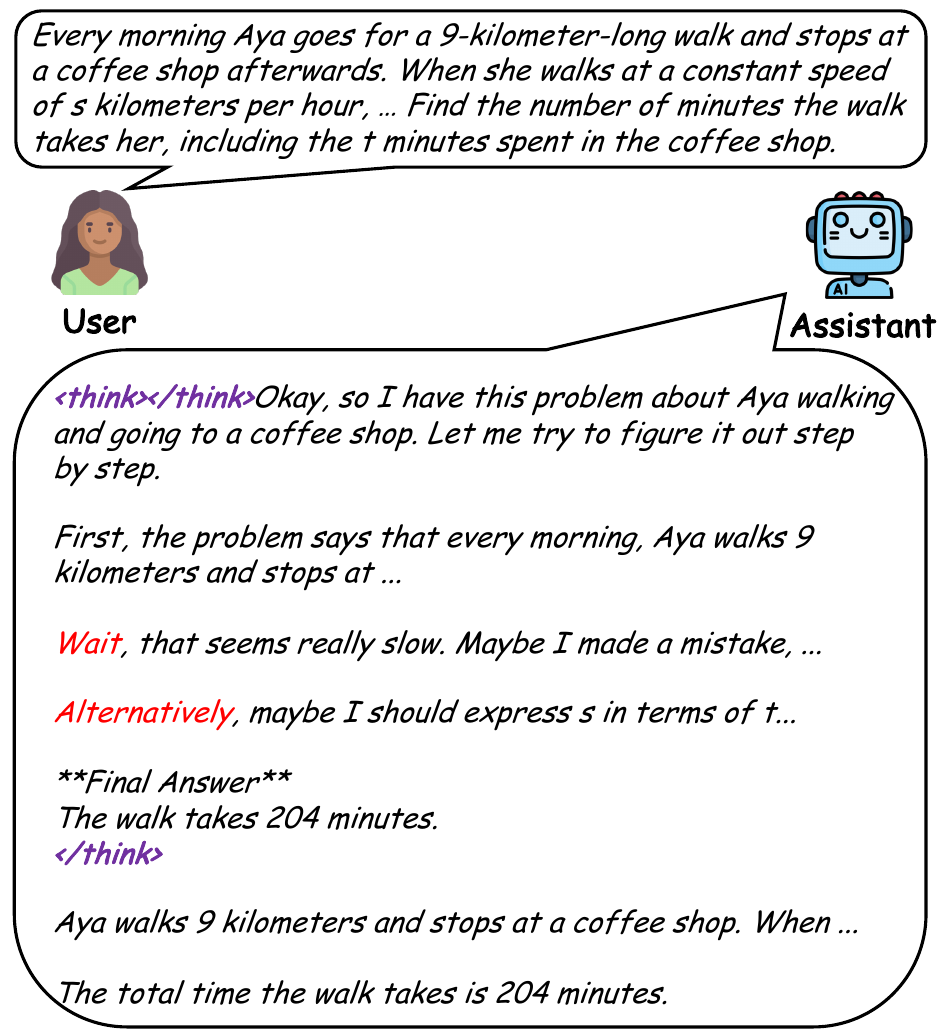}
  \vspace{-5pt}
  \caption{An example of the reward hacking problem occurs when Assistant uses the AutoThink model~\citep{tu2025learning}. The first generated token in the response is $\text{</think>}$ ($\text{<think>}$ is included in the input prompt $x$), indicating that the response is classified as the non-thinking mode. However, the response clearly demonstrates characteristics of the thinking mode, as evidenced by the usage of keywords like "Wait" and "Alternatively" along with the regeneration of the thinking's termination token $\text{</think>}$. The answer $204$ is correct. This response is undeniably part of the thinking mode despite being incorrectly assigned a higher reward associated with the correct answer of the non-thinking mode. This discrepancy results in a clear instance of the reward hacking problem, where the reward allocation does not align with the true mode of the response.
  }
  \label{fig:reward-hacking-example}
  \vspace{-0.6cm}
\end{figure}

As mentioned in \Cref{sec:Preliminary}, \citet{zhang2025adaptthink, tu2025learning} and \citet{fang2025thinkless} determine whether the response is thinking or non-thinking mode based on whether the first token in the response $y$ is $\text{</think>}$ or $\text{<short>}$, respectively. However, the model might generate $\text{</think>}$ or $\text{<short>}$ as the first token, indicating the non-thinking mode. Nonetheless, subsequent tokens conform to the thinking mode, producing a long CoT, as demonstrated in Figure \ref{fig:reward-hacking-example}. Specifically, a reward associated with the non-thinking mode for yielding the correct answer is allocated. However, it is the thinking mode that genuinely derives the correct answer, which is substantiated by the presence of keywords such as "Wait" and "Alternatively", as well as the regeneration of the thinking's termination token $\text{</think>}$. This is a classic reward hacking problem, where the reward is not allocated to the true mode.

Surprisingly, for models that do not consider the reward hacking problem, such as AutoThink~\citep{tu2025learning}, our evaluation of its Stage 1 model performance on AIME24 reveals that responses identified as non-thinking mode exhibit a extremely high average token usage. Specifically, the average token usage for non-thinking mode responses is $10845$, while for thinking mode responses, it is $11976$. This finding highlights that failing to address the reward hacking problem issue could lead to the collapse of the entire training process.

\begin{algorithm*}[!t]
\caption{Thinking-Based Non-Thinking (TNT)}
{
\renewcommand{\arraystretch}{1.0}
\label{alg:TNT}
\textbf{Input:} policy model $\pi_\theta$; dataset $\mathcal{D}$; hyperparameters $\omega, \mathcal{L}^{\emptyset}, \alpha, \beta$
\begin{algorithmic}[1]
    \FOR{$\text{step}=1, \dots, M$} 
        \STATE Sample a batch $\mathcal{D}_b$ from $\mathcal{D}$
        \STATE Sample $K$ responses $\{y_x^k\}_{k=1}^K\sim\pi_{\theta}(\cdot|x)$ for each $x\in \mathcal{D}_b$
        \STATE Split $\{y_x^k\}_{k=1}^K$ into $\mathcal{M}^{\text{T}}_{x} = \{ y_x^j \mid p(y_x^j) = 1\}$ and $\mathcal{M}^{\text{N}}_{x} = \{ y_x^j \mid p(y_x^j) = 0\}$ for each prompt $x$
        \STATE Compute $\mathcal{L}_{x}^{\text{N}}$ via $\mathcal{M}^{\text{T}}_{x}$ (\Cref{eq:token_length_no_thinking}) 
        \STATE Assign a reward for each response $y_x^k$ of each prompt $x$ via $\mathcal{L}_{x}^{\text{N}}$ (\Cref{eq:our reward function}) 
        \STATE Update the policy model $\pi_\theta$ via GPRO and the assigned rewards 
    \ENDFOR
\end{algorithmic}
\textbf{Output:} $\pi_\theta$ 
}
\end{algorithm*}

\section{Our Method}\label{sec:Method}

To mitigate the reward hacking problem that reduces the performance, there are two primary methods: adding SFT, which subsequently raises computational costs, or setting a smaller maximum token usage for the thinking mode than that of the non-thinking mode, 
However, existing research imposes a uniformly maximum token usage for the non-thinking mode across all queries without considering
their varying levels of difficulty, resulting in only limited mitigation of the reward hacking problem (see details in \Cref{subsec:RL-only training method}).

To set the maximum token usage of the non-thinking mode for different queries, we propose \textit{Thinking-Based Non-Thinking} (TNT). The core insight of TNT is to utilize the solution component of the thinking mode responses to determine the maximum token usage for the non-thinking mode. Specifically, LRMs' thinking mode is trained on large-scale datasets, ensuring that the solution component is devoid of any additional thinking. As a result, this solution component closely aligns with the output of the non-thinking mode, which is defined by generating solutions without thinking. Thus, we can appropriately utilize the token usage derived from this solution component to set the maximum token usage for the non-thinking mode.

\subsection{Details of TNT}\label{subsec:Details of TNT}

We now present the details of TNT. The pseudocode for TNT is provided in \Cref{alg:TNT}. 

Consider a reasoning model $\pi_\theta$ and a dataset $\mathcal{D}$. Let $(x, y_x^*) \in \mathcal{D}$ denote the input prompt and golden answer, respectively. We adopt the \textit{ellipsis prompt} from \citet{tu2025learning}, defining $x$ as $[x_1, \ldots, x_n, \text{<think>}, \text{\textbackslash n}, \ldots, \text{\textbackslash n}]$, to enable sampling the non-thinking mode responses without the complex off-policy sampling. For each $(x, y_x^*) \in \mathcal{D}$, we use $y_x^k \sim  \pi_\theta(\cdot|x)$ to denote the $k$-th sampled response of the model $\pi_\theta$ given the prompt $x$. Also, we employ $p(y_x^k)$ to represent the discrimination function that determines whether the mode of the response $y_x^k$ is thinking or non-thinking. This function outputs $1$ if the mode is classified as thinking and $0$ if it is classified as non-thinking. Following \citet{tu2025learning, zhang2025adaptthink, lou2025adacot}, 
$y$ is classified as non-thinking mode if $y$ starts with $\text{</think>}$, otherwise as thinking mode.

The core of TNT is to determine the maximum token usage for the non-thinking mode by leveraging the solution component of responses in the thinking mode. Specifically, for each $(x, y_x^*) \in \mathcal{D}$, we sample $K$ responses, denoted as $y_x^1, y_x^2, \ldots, y_x^k, \ldots, y_x^K$. Let $\mathcal{M}^{\text{T}}_{x} = \{ y_x^j \mid p(y_x^j) = 1\}$ and $\mathcal{M}^{\text{N}}_{x} = \{ y_x^j \mid p(y_x^j) = 0\}$, which correspond to sets of responses in the thinking mode and non-thinking mode given the prompt $x$, respectively. We denote $\mathcal{L}_{x}^{\text{T}}$ and $\mathcal{L}_{x}^{\text{N}}$ as the maximum token usages for the thinking and non-thinking modes given the prompt $x$. We focus on the computation of $\mathcal{L}_{x}^{\text{N}}$, as $\mathcal{L}_{x}^{\text{T}}$ is equal to the training context length and remains unchanged across all prompt $x$. 

For each $y_x^j \in \mathcal{M}^{\text{T}}_{x}$, we evaluate the number of tokens following $\text{</think>}$ in the response $y_x^j = [y_{x,1}^j, \ldots, y_{x,\tau}^j, \text{</think>}, y_{x,\tau+2}^j, \ldots, y_{x,m}^j]$, which is denoted as $h(y_x^j) = m - \tau - 1$. In other words, $h(y_x^j)$ is the length of the token sequence $[y_{x,\tau+2}^j, \ldots, y_{x,m}^j]$. We then define $\mathcal{L}_{x}^{\text{N}}$ as:
\begin{equation}\label{eq:token_length_no_thinking}
\setlength{\abovedisplayskip}{2pt}
\setlength{\belowdisplayskip}{2pt}
\setlength{\abovedisplayshortskip}{2pt}
\setlength{\belowdisplayshortskip}{2pt}
    \mathcal{L}_{x}^{\text{N}} = 
    \begin{cases}
 \omega \frac{\sum_{ y_x^j \in  \mathcal{M}^{\text{T}}_{x} } h(y_x^j)}{|\mathcal{M}^{\text{T}}_{x}|} , & \text{if } \mathcal{M}^{\text{T}}_x \neq \emptyset, \\
\mathcal{L}^{\emptyset},  & \text{if } \mathcal{M}^{\text{T}}_x = \emptyset,
\end{cases}    
\end{equation}
where $\omega \geq 1$ and $\mathcal{L}^{\emptyset}$ are positive constants. The role of $\omega$ in \Cref{eq:token_length_no_thinking} is crucial to reduce the risk of misidentifying reward hacking problem. Formally, we employ average value of $h(y_x^j)$, where $y_x^j \in \mathcal{M}^{\text{T}}_{x}$. If the token usage of a response from $\mathcal{M}^{\text{N}}_{x}$ exceeds the average value by precisely 1, it may be incorrectly identified as exhibiting reward hacking problem. To mitigate this risk, we incorporate a weight $\omega > 1$. This adjustment helps in distinguishing genuine anomalies from minor deviations, thereby addressing the potential misjudgment. In our experiments, we set $\omega=2$. In addition, the parameter $\mathcal{L}^{\emptyset}$ in \Cref{eq:token_length_no_thinking} is introduced because we may not sample the thinking mode responses when we use the on-policy sampling. When this occurs, we are unable to calculate the expression ${\omega \sum_{y_x^j \in \mathcal{M}^{\text{T}}} h(y_x^j)}/{|\mathcal{M}^{\text{T}}_x|}$. Thus, the introduction of the parameter $\mathcal{L}^{\emptyset}$ is necessary. In our experiments, we typically set $\mathcal{L}^{\emptyset}$ to $1000$.

After computing $\mathcal{L}_{x}^{\text{N}}$, we proceed to describe its application in constructing the reward function $R(x, y_x^k, y_x^*, p(y_x^k), \mathcal{L}_{x}^{\text{N}})$. Let $r(y_x^k, y_x^*)$ denote the answer extraction function, which returns $0$ if $y_x^k$ is incorrect and $1$ if $y_x^k$ is correct. Firstly, if the response $y_x^k$ is classified as the thinking mode, e.g., $p(y_x^k)=1$, we use the following reward function:
\begin{equation}\label{eq:our reward function-thinking}
\thinmuskip=0mu
\medmuskip=0mu
\thickmuskip=0mu
\spaceskip=0pt
\setlength{\abovedisplayskip}{2pt}
\setlength{\belowdisplayskip}{2pt}
\setlength{\abovedisplayshortskip}{2pt}
\setlength{\belowdisplayshortskip}{2pt}
\begin{aligned}
     R^{\text{T}}(x, y_x^k, y_x^*,\mathcal{L}_{x}^{\text{N}})  = 
    \begin{cases}
        1, & \text{if } r(y_x^k, y_x^*)=1, \\
        0, & \text{if } r(y_x^k, y_x^*)=0. \\
    \end{cases}
\end{aligned}
\end{equation}
This function indicates that when the response $y^k_x$ is determined to be in thinking mode, a correct answer returns a reward of $1$, while an incorrect response yields $0$. Secondly, if the response $y_x^k$ is classified as the non-thinking mode, e.g., $p(y_x^k)=0$, We use the following reward function that is modified from the one proposed by \citet{tu2025learning}:
\begin{equation}\label{eq:our reward function-non-thinking}
\setlength{\abovedisplayskip}{2pt}
\setlength{\belowdisplayskip}{2pt}
\setlength{\abovedisplayshortskip}{2pt}
\setlength{\belowdisplayshortskip}{2pt}
\begin{aligned}
    & R^{\text{N}}(x, y_x^k, y_x^*, \mathcal{L}_{x}^{\text{N}}) \\ &  =
    \begin{cases}
        2, & \text{if } r(y_x^k, y_x^*)=1 \land |y_x^k| \leq \mathcal{L}_{x}^{\text{N}},  \\
        -1, & \text{if } r(y_x^k, y_x^*)=0 \land |y_x^k| \leq \mathcal{L}_{x}^{\text{N}},  \\
        -2, & \text{if }  |y_x^k| > \mathcal{L}_{x}^{\text{N}},  \\
    \end{cases}
\end{aligned}
\end{equation}
where $|y_x^k|$ represents the length of $y_x^k$. This function implies that if $|y_x^k|$ is less than or equal to $\mathcal{L}_{x}^{\text{N}}$, indicating no reward hacking problem, we directly employ the naive reward function as presented in \citet{tu2025learning}. 
This incentivizes using the non-thinking mode when both non-thinking and thinking modes can yield correct responses, thereby allowing for automatic mode selection based on query complexity. Conversely, when $|y_x^k|$ exceeds $\mathcal{L}_{x}^{\text{N}}$, reward hacking problem is indicated. By setting the reward as $-2$ for this scenario, we ensure that both incorrect and correct answers are assigned lower rewards compared to scenarios without reward hacking problem, thereby minimizing the reward hacking problem behavior.

By integrating Equations (\ref{eq:our reward function-thinking}) and (\ref{eq:our reward function-non-thinking}), which correspond to the reward functions for thinking and non-thinking mode responses, respectively, the ultimate reward function can be expressed as follows:
\begin{equation}\label{eq:our reward function}
\setlength{\abovedisplayskip}{2pt}
\setlength{\belowdisplayskip}{2pt}
\setlength{\abovedisplayshortskip}{2pt}
\setlength{\belowdisplayshortskip}{2pt}
\begin{aligned}
    & R(x, y_x^k, y_x^*, p(y_x^k),\mathcal{L}_{x}^{\text{N}})  \\ & 
    =
    \begin{cases}
        R^{\text{T}}(x, y_x^k, y_x^*,\mathcal{L}_{x}^{\text{N}}), & \text{if } p(y_x^k)=1, \\
        R^{\text{N}}(x, y_x^k, y_x^*,\mathcal{L}_{x}^{\text{N}}), & \text{if } p(y_x^k)=0. \\
    \end{cases}
\end{aligned}
\end{equation}

Finally, according to the dataset $\mathcal{D}$ and the reward function $R(x, y_x^k, y_x^*, p(y_x^k),\mathcal{L}_{x}^{\text{N}})$ defined in \Cref{eq:our reward function}, we employ the GRPO algorithm with a token-level policy gradient loss for training~\citep{shao2024deepseekmath}, as defined by: 
\begin{equation}\label{eq:grpo}
\thinmuskip=0mu
\medmuskip=0mu
\thickmuskip=0mu
\spaceskip=0pt
\setlength{\abovedisplayskip}{0pt}
\setlength{\belowdisplayskip}{0pt}
\setlength{\abovedisplayshortskip}{0pt}
\setlength{\belowdisplayshortskip}{0pt}
\begin{aligned}
& \mathcal{J}(\theta) = 
\mathbb{E}_{(x, y_x^*) \sim \mathcal{D},\; \{y_x^k\}_{k=1}^K \sim \pi_{\theta_{\text{old}}}(\cdot \mid x)} 
 \Big[
\frac{1}{\sum_{k=1}^{K} |y_x^k|}   
\sum_{k=1}^{K}  \\ & \sum_{t=1}^{|y_x^k|}  
\min \Big(
r_{i,t}(\theta) \hat{A}_{i,t}, \nonumber 
 \text{clip}\left(r_{i,t}(\theta), 1 - \varepsilon, 1 + \varepsilon \right) \hat{A}_{i,t}
\Big)
\Big],
\end{aligned}
\end{equation}
where \( r_{i,t}(\theta) \) is the token-level importance weight defined as the ratio between the new and old token probabilities, and \( \hat{A}_{i,t} \) represents the token-level advantage estimated through the reward function $ R(x, y_x^k, y_x^*, p(y_x^k),\mathcal{L}_{x}^{\text{N}})$ defined in \Cref{eq:our reward function}. The overall loss is normalized by the total number of tokens across all sampled trajectories.

\subsection{Discussions of TNT}\label{subsec:Discussions of TNT}

The core of TNT focuses on determining the maximum token usage for the non-thinking mode, which enables TNT to be compatible with RL algorithms. This implies that, while GRPO is used in the design of TNT, it can be replaced with other advanced GRPO variants, such as Dr. GRPO~\citep{liu2025_drGRPO},  DAPO~\citep{yu2025dapo}, and GSPO~\citep{zheng2025_gspo}, or the classic PPO~\citep{schulman2017proximal}. Also, as discussed in \Cref{subsec:Details of TNT}, TNT can also be effectively combined with the off-policy sampling, as did in \citet{zhang2025adaptthink} and \citet{jiang2025_LHRMs}. Additionally, techniques from other research on hybrid reasoning models can be applied. For instance, we can assign different weights to the control and response tokens in the loss function~\citep{fang2025thinkless,lou2025adacot, zhang2025adaptthink}. Furthermore, the two methods employed by \citet{tu2025learning}—Batch-Level Reward Balancing and Length-Aware Reward—can also be utilized to respectively alleviate model collapse and further reduce token usage. Moreover, the reference model reward introduced in \citet{zhang2025adaptthink} can also be incorporated to provide additional signals during the training.

\section{Experiments}\label{sec:Experiments}
\subsection{Experimental Setups}\label{subsec:Experimental Setups}

\begin{table*}[!t]
\centering
\small
\newcolumntype{Y}{>{\centering\arraybackslash}X}
\setlength{\tabcolsep}{1pt} 
\renewcommand{\arraystretch}{1.2}

\begin{tabularx}{\linewidth}{l l *{13}{Y}} 
\toprule
& \multicolumn{1}{c}{\multirow{2}{*}{\textbf{Models}}} & \multicolumn{6}{c}{\textbf{Accuracy (\%) $\uparrow$}} & \multicolumn{6}{c}{\textbf{Token Usage $\downarrow$}} & \multirow{2}{*}{\textbf{TE}} \\
\cmidrule(lr){3-8} \cmidrule(lr){9-14}
& & \scriptsize AIME24 & \scriptsize AIME25 & \scriptsize Minerva & \scriptsize AMC23 & \scriptsize Olym. & \textbf{\scriptsize AVG} & \scriptsize AIME24 & \scriptsize AIME25 & \scriptsize Minerva & \scriptsize AMC23 & \scriptsize Olym. & \textbf{\scriptsize AVG} & \\
\midrule

\rowcolor{gray!8}
& Base Model & 28.6 & 24.6 & 26.2 & 62.1 & 43.6 & 37.0 & 16865 & 16464 & 7490 & 11050 & 11808 & 12736 & 0.33 \\
\rowcolor{cyan!6}
& Thinkless & 28.4 & 24.2 & 26.6 & 58.7 & 42.5 & 36.1 & 11207 & 10664 & 5463 & 7153 & 7379 & 8373 & 0.39 \\
\rowcolor{cyan!6}
& Adapt-0 & 31.9 & 25.0 & 27.3 & \textbf{68.2} & 45.4 & 39.6 & 10140 & 10039 & 4145 & 5661 & 6796 & 7356 & 0.46 \\
\rowcolor{cyan!6}
& Adapt-2 & 32.5 & 24.2 & 23.3 & \underline{66.4} & 43.9 & 38.1 & 9087 & 8367 & \underline{2283} & 4254 & 5252 & \underline{5849} & \underline{0.50} \\
\rowcolor{cyan!6}
& Adapt-5 & 29.5 & 22.9 & 25.9 & 60.8 & 41.2 & 36.1 & \textbf{8105} & \textbf{8038} & \textbf{1661} & \textbf{3475} & \textbf{4240} & \textbf{5104} & \underline{0.50} \\
\rowcolor{cyan!6}
& Auto-S1 & 27.1 & 20.8 & 22.3 & 60.7 & 41.6 & 34.5 & 11701 & 11317 & 6553 & \underline{3822} & 7308 & 8140 & 0.38 \\
\rowcolor{cyan!6}
& Auto-S2 & \underline{35.0} & \textbf{26.3} & \underline{28.0} & 65.1 & \textbf{47.0} & \underline{40.3} & 11858 & 11186 & 5861 & 7178 & 7768 & 8770 & 0.43 \\
\rowcolor{cyan!6}
& Auto-S3 & 34.1 & 25.0 & \textbf{28.7} & 65.8 & 46.2 & 40.0 & 9742 & 8625 & 3172 & 4875 & 5345 & 6352 & \underline{0.50} \\
\rowcolor{orange!10} 
& \textit{\textbf{TNT (Ours)}} & \textbf{37.7} & \underline{26.1} & \textbf{28.7} & 65.9 & \underline{46.4} & \textbf{41.0} & \underline{8537} & \underline{8252} & 3150 & 4508 & \underline{5020} & 5893 & \textbf{0.53} \\
\bottomrule
\end{tabularx}
\vspace{-0.2cm}
\caption{Comparison of accuracy, token usage, and TE on mathematical benchmarks across hybrid reasoning models when the base model is DeepSeek-R1-Distill-Qwen-1.5B. The best and second results are bolded and underlined, respectively. \textcolor{gray}{Gray} represents the base model, \textcolor{cyan}{cyan} represents baseline hybrid reasoning models training methods, and \textcolor{orange}{orange} represents TNT model. We will use the same color settings below as well.}
\vspace{-0.2cm}
\label{tab:main}
\end{table*}

\begin{table*}[!t]
\centering
\small
\newcolumntype{Y}{>{\centering\arraybackslash}X}
\setlength{\tabcolsep}{1pt} 
\renewcommand{\arraystretch}{1.2}

\begin{tabularx}{\linewidth}{l l *{12}{Y}} 
\toprule
& \multicolumn{1}{c}{\multirow{2}{*}{\textbf{Models}}} & \multicolumn{6}{c}{\textbf{Non-Thinking Mode Tokens}} & \multicolumn{6}{c}{\textbf{Thinking Mode Tokens}} \\
\cmidrule(lr){3-8} \cmidrule(lr){9-14} 
& & \scriptsize AIME24 & \scriptsize AIME25 & \scriptsize Minerva & \scriptsize AMC23 & \scriptsize Olym. & \textbf{\scriptsize AVG} & \scriptsize AIME24 & \scriptsize AIME25 & \scriptsize Minerva & \scriptsize AMC23 & \scriptsize Olym. & \textbf{\scriptsize AVG} \\
\midrule

\rowcolor{gray!8}
& Base Model & -- & -- & -- & -- & -- & -- & 16865 & 16464 & 7490 & 11050 & 11808 & 12736 \\

\rowcolor{cyan!6}
& Thinkless & -- & -- & 372 & 367 & 467 & 402 & 11439 & 10670 & 5535 & 7510 & 8649 & 8761 \\
\rowcolor{cyan!6}
& Adapt-0 & 5468 & 2283 & 1059 & 3507 & 3912 & 3246 & 10536 & 11216 & 4448 & 6513 & 8759 & 8294 \\
\rowcolor{cyan!6}
& Adapt-2 & 6549 & 4724 & 1586 & 3034 & 4624 & 4103 & 11050 & 10258 & 3807 & 5957 & 8206 & 7856 \\
\rowcolor{cyan!6}
& Adapt-5 & 4076 & 2556 & 961 & 1499 & 2314 & 2281 & 11137 & 11041 & 5558 & 8514 & 9268 & 9104 \\
\rowcolor{cyan!6}
& Auto-S1 & 10845 & 8858 & 1893 & 3980 & 3806 & 5876 & 11976 & 12223 & 4212 & 10114 & 10033 & 9712 \\
\rowcolor{cyan!6}
& Auto-S2 & 7870 & 6507 & 4995 & 5024 & 5521 & 5983 & 12777 & 12066 & 6815 & 11060 & 11081 & 10760 \\
\rowcolor{cyan!6}
& Auto-S3 & 4537 & 1890 & 1559 & 2107 & 2759 & 2570 & 10655 & 8601 & 3729 & 6950 & 7130 & 7413 \\

\rowcolor{orange!10} 
& \textit{\textbf{TNT (Ours)}} & 995 & 795 & 601 & 859 & 937 & 837 & 8633 & 8325 & 3475 & 7159 & 6086 & 6736 \\

\bottomrule
\end{tabularx}
\vspace{-0.3cm}
\caption{Comparison of non-thinking and thinking mode token usage on mathematical benchmarks across hybrid reasoning models when the base model is DeepSeek-R1-Distill-Qwen-1.5B.}
\vspace{-0.4cm}
\label{tab:two mode token_usage}
\end{table*}

\begin{table}[t]
\centering
\small
\newcolumntype{Y}{>{\centering\arraybackslash}X}
\setlength{\tabcolsep}{1pt} 
\renewcommand{\arraystretch}{1.2}

\begin{tabularx}{\linewidth}{l *{5}{Y}} 
\toprule
\multicolumn{1}{c}{\multirow{2}{*}{\textbf{Models}}} & \multicolumn{5}{c}{\textbf{No-Thinking Mode Ratio (\%)}} \\
\cmidrule(lr){2-6}
& \scriptsize AIME24 & \scriptsize AIME25 & \scriptsize Minerva & \scriptsize AMC23 & \scriptsize Olym. \\
\midrule
\rowcolor{gray!8}
Base Model & 0.0 & 0.0 & 0.0 & 0.0 & 0.0 \\
\rowcolor{cyan!6}
Thinkless & 0.0 & 0.0 & 1.4 & 5.0 & 2.8 \\
\rowcolor{cyan!6}
Adapt-0 & 11.3 & 9.6 & 9.0 & 28.3 & 40.5 \\
\rowcolor{cyan!6}
Adapt-2 & 42.9 & 35.4 & 68.6 & 58.3 & 71.3 \\
\rowcolor{cyan!6}
Adapt-5 & 44.2 & 36.7 & 84.7 & 71.8 & 72.3 \\
\rowcolor{cyan!6}
Auto-S1 & 22.8 & 25.4 & 16.8 & 58.1 & 37.3 \\
\rowcolor{cyan!6}
Auto-S2 & 19.0 & 17.9 & 52.4 & 64.3 & 59.6 \\
\rowcolor{cyan!6}
Auto-S3 & 13.3 & 10.2 & 29.0 & 47.6 & 46.8 \\
\rowcolor{orange!10}
\textit{\textbf{TNT (Ours)}} & 1.7 & 0.8 & 11.3 & 29.4 & 20.7 \\

\bottomrule
\end{tabularx}
\vspace{-0.2cm}
\caption{Non-thinking mode ratio on mathematical benchmarks across hybrid reasoning models when the base model is DeepSeek-R1-Distill-Qwen-1.5B.}
\vspace{-0.5cm}
\label{tab:No-Thinking Mode Ratio}
\end{table}

\textbf{Datasets and Models.} Following previous hybrid reasoning model training methods~\citep{fang2025thinkless,zhang2025adaptthink,tu2025learning}, we utilize DeepScaleR~\citep{luo2025DeepScaleR} as the training dataset, which consists of 40,000 mathematical problems with varying levels of difficulty. Following previous hybrid reasoning model training methods, we select DeepSeek-R1-Distill-Qwen-1.5B~\citep{guo2025deepseek} as the base model in this section, given that it serves as the only base model shared across all previous hybrid reasoning model training methods. Besides, we show more details for the scenario when the base models are DeepScaleR-1.5B~\citep{luo2025DeepScaleR} or DeepSeek-R1-Distill-Qwen-7B~\citep{guo2025deepseek} in Finding \ref{more models} and \Cref{subsec:deepscaler and 7B}. Our evaluation is performed on five standard mathematical benchmarks: AIME24, AIME25, Minerva, AMC23, and Olympiad. We do not conduct tests on the GSM8K~\citep{cobbe2021-gsm8k} and MATH~\citep{lightman2023lets-math500}, as both are too simplistic to adequately reflect the model's performance. 
During evaluation, all models use a 32K context window. For evaluation metrics, we consider accuracy and response token usage.

\textbf{Training Details.} All experiments are conducted on a single node with 8 H800 GPUs. All experiments are implemented using the verl framework~\citep{sheng2025hybridflow}. The batch size and training context length are set to 64 and 16K, respectively. We selected the checkpoint at step 600 when the base model is DeepSeek-R1-Distill-Qwen-1.5B, as it observed convergence and yielded the optimal performance. 
We set $\omega=2$ and $\mathcal{L}^{\emptyset}=1000$, as mentioned in \Cref{subsec:Details of TNT}.

\textbf{Baselines.} We mainly evaluate TNT against representative methods for hybrid reasoning model training method, such as Thinkless~\citep{fang2025thinkless}, AdaptThink~\citep{zhang2025adaptthink}, and AutoThink~\citep{tu2025learning}. They are selected based on the public availability of their trained models, as well as their base models consistent with our experiments.  Furthermore, regarding specific configurations: for Thinkless, we exclusively analyze its RL model due to the suboptimal performance of its SFT model; for AdaptThink, we evaluate configurations with $\delta \in \{0.0, 0.02, 0.05\}$, noting that $\delta=0.05$ serves as the default; and for AutoThink, we assess performance across its three distinct training stages (Stage 1 to Stage 3).

\subsection{Experimental Results}\label{subsec:Experimental Results}

\findingbox{TNT enhances accuracy and simultaneously significantly reduces token usage.}

Table \ref{tab:main} presents the accuracy and token usage of various models when the base model is DeepSeek-R1-Distill-Qwen-1.5B across the tested datasets. To evaluate the trade-off between accuracy and conciseness, we adopt the Token Efficiency (TE) metric from \citet{hong2025reconsidering}, defined as $A/\sqrt{L}$, where $A$ and $L$ represent accuracy and token usage, respectively. The average accuracy and token usage across all datasets are used to compute this metric. 
Additionally, the token usage on each benchmark for both the thinking mode and non-thinking mode responses are shown in \Cref{tab:two mode token_usage}.

Firstly, compared to the base model, TNT reduces the average response token usage by $46.2\%$ and enhances the average accuracy by $4.1\%$. Secondly, TNT surpasses Thinkless across all datasets in terms of both accuracy and token usage. On the AIME24 dataset, TNT utilizes approximately $80\%$ of the tokens used by Thinkless while exceeding their accuracy by over $10\%$. Notably, TNT and Thinkless employed the same RL dataset--DeepScaleR. Thirdly, compared to AdaptThink, while TNT primarily focuses on enhancing accuracy, it surpasses three AdaptThink models trained with different $\delta$ by evaluating the trade-off between accuracy and conciseness through the TE. Lastly, TNT consistently outperforms the final output model of AutoThink (AutoThink-Stage 3) across all benchmarks in both accuracy and token usage. Notably, TNT achieves this without the CoT Compression method used in AutoThink. While incorporating CoT Compression methods may further reduce TNT's token usage, we opt not to pursue this as our focus remains on the hybrid reasoning models rather than CoT Compression methods.

In fact, we also examine the variations in accuracy and token usage on both the training and testing datasets during the training process. See details in Appendix \ref{subsec:deepscaler and 7B}.

\begin{figure*}[t]
  \centering
  \includegraphics[width=0.97\textwidth]{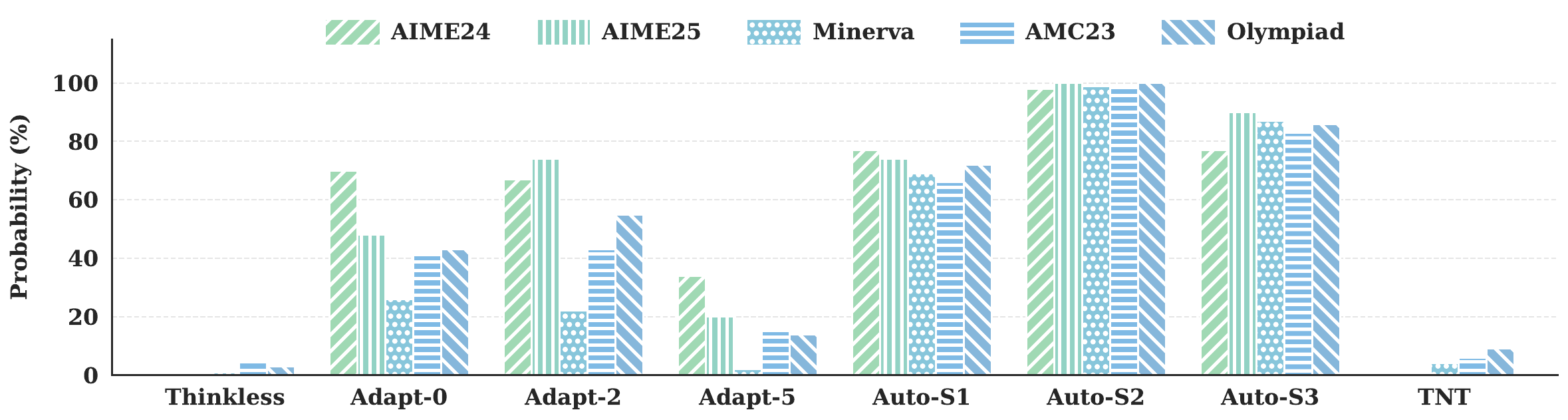}
  \vspace{-10pt}
  \caption{Probability of thinking-related verbs appearing in non-thinking mode responses across models with base model DeepSeek-R1-Distill-Qwen-1.5B on mathematical benchmarks.}
  \label{fig:Probability_of_Reasoning-Related_Verbs}
  \vspace{-0.4cm}
\end{figure*}

\findingbox{TNT learns to autonomously select between the non-thinking mode and thinking mode based on the difficulty of the query.}

We also investigate the relationship between thinking behavior and the inherent difficulty of tasks. 
In \Cref{tab:No-Thinking Mode Ratio}, we report the ratio of responses classified as belonging to the non-thinking mode across different models.
By considering the results in \Cref{tab:main}, we observe a negative correlation between the non-thinking rate and task difficulty. Specifically, accuracy serves as a proxy for dataset difficulty. We find that on more challenging datasets, TNT tends to invoke explicit thinking more frequently. In addition, the results in Appendix \ref{subsec:deepscaler and 7B} (\Cref{fig:training dynamics}) illustrate the variation of the non-thinking mode ratio on both the training and testing datasets during the training process. We observe that while such ratio on the training set exhibits a U-shaped pattern similar to Thinkless, such ratio on the testing dataset consistently remains at a low level. This observation indicates that TNT does not exhibit overfitting.

\findingbox{TNT effectively mitigates the reward hacking problem.}

We analyze the probability of thinking-related verbs such as "Wait," "Alternatively," and "Double-Check" appearing in responses classified under the non-thinking mode across different models, as shown in Figure~\ref{fig:Probability_of_Reasoning-Related_Verbs}. This probability reflects the likelihood of reward hacking problem in the model responses, as the presence of these words indicates actual thinking by the model. Our findings show that Thinkless, which utilizes SFT, exhibits the lowest probability, followed by our TNT model. Additionally, we find that the uniform maximum token usage for all queries in the non-thinking mode used in AdaptThink significantly increases the likelihood of reward hacking problem compared to TNT. Also, AutoThink, which does not consider the reward hacking problem, shows the highest probability. Furthermore, results in \Cref{tab:two mode token_usage} corroborate that TNT effectively mitigates the reward hacking problem. Specifically, in terms of the token usage for responses in non-thinking mode, Thinkless has the lowest value, with TNT as the second lowest. 

Although Thinkless addresses the reward hacking problem more effectively, it relies on SFT, which results in significant computational costs. Specifically, the SFT dataset used by Thinkless is approximately 2000K in size, making it about 50 times larger than the RL dataset, DeepScaleR, which is employed by Thinkless, AdaptThink, AutoThink, and ours.
Moreover, constructing the SFT dataset incurs substantial computational overhead. Thinkless utilizes DeepSeek-R1-671B~\citep{guo2025deepseek} to generate the SFT dataset, a process that is computationally intensive. In addition, SFT leads to significant performance degradation, which is why we do not compare against Thinkless's SFT model. Precisely, the accuracy of Thinkless's SFT model on AIME24 is merely around $10\%$!

\findingbox[more models]{TNT becomes more pronounced on better base models.}
To verify the scalability of TNT, we conducted experiments on DeepScaleR-1.5B and DeepSeek-R1-Distill-Qwen-7B. Interestingly, TNT attains TE scores of 0.70 and 0.79, respectively, significantly surpassing the runners-up (0.54 and 0.67). Our results demonstrate that the advantage of TNT becomes more pronounced as the capability of the base model increases. See details in Appendix \ref{subsec:deepscaler and 7B}.

\findingbox{TNT exhibits better performance compared with CoT compression methods.}
Besides, we compared our accuracy and token usage against CoT compression methods~\citep{thinkprune, reasoningefficient}. Also, we exclusively evaluate open-source checkpoints.
While CoT compression methods typically show a drop in accuracy as compression intensifies, TNT maintains or even improves the model's accuracy, achieving the highest TE scores. This establishes TNT not merely as a compression tool, but as a reasoning enhancement framework that fundamentally reorganizes how models allocate computational resources.
See details in Appendix \ref{subsec:cot compression}.

\findingbox{TNT exhibits excellent performance in Out of Distribution settings.}
We also evaluate the performance of TNT in out-of-distribution (OOD) tasks. The experimental results show that TNT exhibits the best TE across all tested models. See details in Appendix \ref{subsec:Experiments in OOD}.

\findingbox{Reward hacking problem appears without our reward function.}
Finally, we conduct the ablation study. 
Specifically, we test under the scenario where the components of our reward function specifically related to reward hacking problem are removed.
Consequently, reward hacking problem occurs. See more details in Appendix \ref{subsec:Ablation Study}.

\section{Conclusion}\label{sec:Conclusion}

We explore the reward hacking problem in training hybrid reasoning models via RL, where correct answers from the no-thinking mode are rewarded more highly than those from the thinking mode. To mitigate this problem, we propose TNT, which leverages the solution component of the thinking mode to adaptively determine the maximum token usage of the no-thinking mode for different queries. This improves the poor detection of reward hacking in previous approaches that set a uniform maximum token usage across all queries, thereby effectively mitigates the reward hacking problem and achieves superior performance. Experiments show that TNT significantly improves accuracy and token efficiency while minimizing reward hacking.

\section*{Limitations}\label{sec:Limitations}

While TNT shows promising adaptive thinking capabilities, it encounters several limitations. First, similar to most prior open-source studies, we confine our model training to mathematical datasets. These datasets are preferred because they are easily accessible and can offer accurate, verifiable rewards. Furthermore, as discussed in \Cref{subsec:Discussions of TNT}, while TNT has the potential to be integrated with various techniques, our current study lacks sufficient testing due to computational resource constraints. As such, we have not explored whether combining TNT with these techniques can enhance performance.

\section*{Acknowledgments}

This work is funded by Nanjing University China Mobile Communications Group Co.,Ltd. Joint Institute, National Natural Science Foundation of China (62192783, 62276128, 62406111), Jiangsu Natural Science Foundation (BK20243051), Jiangsu Science and Technology Major Project (BG2024031), the Fundamental Research Funds for the Central Universities(14380128, KG202514), the Fundamental and Interdisciplinary Disciplines Breakthrough Plan of the Ministry of Education of China (No. JYB2025XDXM118), the Collaborative Innovation Center of Novel Software Technology and Industrialization, and Shanghai Artificial Intelligence Laboratory.

\bibliography{custom}

@article{guo2025deepseek,
  title={Deepseek-r1: Incentivizing reasoning capability in llms via reinforcement learning},
  author={Guo, Daya and Yang, Dejian and Zhang, Haowei and Song, Junxiao and Zhang, Ruoyu and Xu, Runxin and Zhu, Qihao and Ma, Shirong and Wang, Peiyi and Bi, Xiao and others},
  journal={arXiv preprint arXiv:2501.12948},
  year={2025}
}

@article{jaech2024openai,
  title={Openai o1 system card},
  author={Jaech, Aaron and Kalai, Adam and Lerer, Adam and Richardson, Adam and El-Kishky, Ahmed and Low, Aiden and Helyar, Alec and Madry, Aleksander and Beutel, Alex and Carney, Alex and others},
  journal={arXiv preprint arXiv:2412.16720},
  year={2024}
}

@article{shao2024deepseekmath,
  title={Deepseekmath: Pushing the limits of mathematical reasoning in open language models},
  author={Shao, Zhihong and Wang, Peiyi and Zhu, Qihao and Xu, Runxin and Song, Junxiao and Bi, Xiao and Zhang, Haowei and Zhang, Mingchuan and Li, YK and Wu, Yang and others},
  journal={arXiv preprint arXiv:2402.03300},
  year={2024}
}

@article{sui2025stop,
  title={Stop overthinking: A survey on efficient reasoning for large language models},
  author={Sui, Yang and Chuang, Yu-Neng and Wang, Guanchu and Zhang, Jiamu and Zhang, Tianyi and Yuan, Jiayi and Liu, Hongyi and Wen, Andrew and Zhong, Shaochen and Chen, Hanjie and others},
  journal={arXiv preprint arXiv:2503.16419},
  year={2025}
}

@article{xu2025towards,
  title={Towards large reasoning models: A survey of reinforced reasoning with large language models},
  author={Xu, Fengli and Hao, Qianyue and Zong, Zefang and Wang, Jingwei and Zhang, Yunke and Wang, Jingyi and Lan, Xiaochong and Gong, Jiahui and Ouyang, Tianjian and Meng, Fanjin and others},
  journal={arXiv preprint arXiv:2501.09686},
  year={2025}
}

@article{wei2022chain,
  title={Chain-of-thought prompting elicits reasoning in large language models},
  author={Wei, Jason and Wang, Xuezhi and Schuurmans, Dale and Bosma, Maarten and Xia, Fei and Chi, Ed and Le, Quoc V and Zhou, Denny and others},
  journal={Advances in neural information processing systems},
  volume={35},
  year={2022}
}

@article{qu2025survey,
  title={A survey of efficient reasoning for large reasoning models: Language, multimodality, and beyond},
  author={Qu, Xiaoye and Li, Yafu and Su, Zhaochen and Sun, Weigao and Yan, Jianhao and Liu, Dongrui and Cui, Ganqu and Liu, Daizong and Liang, Shuxian and He, Junxian and others},
  journal={arXiv preprint arXiv:2503.21614},
  year={2025}
}

@article{li2025system,
  title={From system 1 to system 2: A survey of reasoning large language models},
  author={Li, Zhong-Zhi and Zhang, Duzhen and Zhang, Ming-Liang and Zhang, Jiaxin and Liu, Zengyan and Yao, Yuxuan and Xu, Haotian and Zheng, Junhao and Wang, Pei-Jie and Chen, Xiuyi and others},
  journal={arXiv preprint arXiv:2502.17419},
  year={2025}
}

@article{chen2024not,
  title={Do not think that much for 2+ 3=? on the overthinking of o1-like llms},
  author={Chen, Xingyu and Xu, Jiahao and Liang, Tian and He, Zhiwei and Pang, Jianhui and Yu, Dian and Song, Linfeng and Liu, Qiuzhi and Zhou, Mengfei and Zhang, Zhuosheng and others},
  journal={arXiv preprint arXiv:2412.21187},
  year={2024}
}

@article{zhang2025adaptthink,
  title={Adaptthink: Reasoning models can learn when to think},
  author={Zhang, Jiajie and Lin, Nianyi and Hou, Lei and Feng, Ling and Li, Juanzi},
  journal={arXiv preprint arXiv:2505.13417},
  year={2025}
}

@article{fang2025thinkless,
  title={Thinkless: Llm learns when to think},
  author={Fang, Gongfan and Ma, Xinyin and Wang, Xinchao},
  journal={arXiv preprint arXiv:2505.13379},
  year={2025}
}

@article{tu2025learning,
  title={Learning When to Think: Shaping Adaptive Reasoning in R1-Style Models via Multi-Stage RL},
  author={Tu, Songjun and Lin, Jiahao and Zhang, Qichao and Tian, Xiangyu and Li, Linjing and Lan, Xiangyuan and Zhao, Dongbin},
  journal={arXiv preprint arXiv:2505.10832},
  year={2025}
}

@article{lou2025adacot,
  title={AdaCoT: Pareto-Optimal Adaptive Chain-of-Thought Triggering via Reinforcement Learning},
  author={Lou, Chenwei and Sun, Zewei and Liang, Xinnian and Qu, Meng and Shen, Wei and Wang, Wenqi and Li, Yuntao and Yang, Qingping and Wu, Shuangzhi},
  journal={arXiv preprint arXiv:2505.11896},
  year={2025}
}

@article{jiang2025_LHRMs,
  title={Think only when you need with large hybrid-reasoning models},
  author={Jiang, Lingjie and Wu, Xun and Huang, Shaohan and Dong, Qingxiu and Chi, Zewen and Dong, Li and Zhang, Xingxing and Lv, Tengchao and Cui, Lei and Wei, Furu},
  journal={arXiv preprint arXiv:2505.14631},
  year={2025}
}

@article{chen2025pangu,
  title={Pangu Embedded: An Efficient Dual-system LLM Reasoner with Metacognition},
  author={Chen, Hanting and Wang, Yasheng and Han, Kai and Li, Dong and Li, Lin and Bi, Zhenni and Li, Jinpeng and Wang, Haoyu and Mi, Fei and Zhu, Mingjian and others},
  journal={arXiv preprint arXiv:2505.22375},
  year={2025}
}

@article{zhan2025kat,
  title={Kat-v1: Kwai-autothink technical report},
  author={Zhan, Zizheng and Deng, Ken and Tang, Huaixi and Xiang, Wen and Wu, Kun and Li, Weihao and Zhu, Wenqiang and Xu, Jingxuan and Huang, Lecheng and Feng, Zongxian and others},
  journal={arXiv preprint arXiv:2507.08297},
  year={2025}
}

@article{cobbe2021-gsm8k,
  title={Training verifiers to solve math word problems},
  author={Cobbe, Karl and Kosaraju, Vineet and Bavarian, Mohammad and Chen, Mark and Jun, Heewoo and Kaiser, Lukasz and Plappert, Matthias and Tworek, Jerry and Hilton, Jacob and Nakano, Reiichiro and others},
  journal={arXiv preprint arXiv:2110.14168},
  year={2021}
}

@article{yu2025dapo,
  title={Dapo: An open-source llm reinforcement learning system at scale},
  author={Yu, Qiying and Zhang, Zheng and Zhu, Ruofei and Yuan, Yufeng and Zuo, Xiaochen and Yue, Yu and Dai, Weinan and Fan, Tiantian and Liu, Gaohong and Liu, Lingjun and others},
  journal={arXiv preprint arXiv:2503.14476},
  year={2025}
}

@article{liu2025_drGRPO,
  title={Understanding r1-zero-like training: A critical perspective},
  author={Liu, Zichen and Chen, Changyu and Li, Wenjun and Qi, Penghui and Pang, Tianyu and Du, Chao and Lee, Wee Sun and Lin, Min},
  journal={arXiv preprint arXiv:2503.20783},
  year={2025}
}

@article{zheng2025_gspo,
  title={Group sequence policy optimization},
  author={Zheng, Chujie and Liu, Shixuan and Li, Mingze and Chen, Xiong-Hui and Yu, Bowen and Gao, Chang and Dang, Kai and Liu, Yuqiong and Men, Rui and Yang, An and others},
  journal={arXiv preprint arXiv:2507.18071},
  year={2025}
}

@article{schulman2017proximal,
  title={Proximal policy optimization algorithms},
  author={Schulman, John and Wolski, Filip and Dhariwal, Prafulla and Radford, Alec and Klimov, Oleg},
  journal={arXiv preprint arXiv:1707.06347},
  year={2017}
}

@article{luo2025deepscaler,
  title={Deepscaler: Surpassing o1-preview with a 1.5 b model by scaling rl},
  author={Luo, Michael and Tan, Sijun and Wong, Justin and Shi, Xiaoxiang and Tang, William Y and Roongta, Manan and Cai, Colin and Luo, Jeffrey and Zhang, Tianjun and Li, Li Erran and others},
  journal={Notion Blog},
  year={2025}
}

@inproceedings{sheng2025hybridflow,
  title={Hybridflow: A flexible and efficient rlhf framework},
  author={Sheng, Guangming and Zhang, Chi and Ye, Zilingfeng and Wu, Xibin and Zhang, Wang and Zhang, Ru and Peng, Yanghua and Lin, Haibin and Wu, Chuan},
  booktitle={Proceedings of the Twentieth European Conference on Computer Systems},
  pages={1279--1297},
  year={2025}
}

@article{lightman2023lets-math500,
      title={Let's Verify Step by Step}, 
      author={Lightman, Hunter and Kosaraju, Vineet and Burda, Yura and Edwards, Harri and Baker, Bowen and Lee, Teddy and Leike, Jan and Schulman, John and Sutskever, Ilya and Cobbe, Karl},
      journal={arXiv preprint arXiv:2305.20050},
      year={2023}
}

@article{hong2025reconsidering,
  title={Reconsidering overthinking: Penalizing internal and external redundancy in cot reasoning},
  author={Hong, Jialiang and Zhen, Taihang and Chen, Kai and Liu, Jiaheng and Zhu, Wenpeng and Huo, Jing and Gao, Yang and Wang, Depeng and Wan, Haitao and Yang, Xi and others},
  journal={arXiv preprint arXiv:2508.02178},
  year={2025}
}

@article{skalse2022defining,
  title={Defining and characterizing reward gaming},
  author={Skalse, Joar and Howe, Nikolaus and Krasheninnikov, Dmitrii and Krueger, David},
  journal={Advances in Neural Information Processing Systems},
  volume={35},
  pages={9460--9471},
  year={2022}
}

@article{pan2402feedback,
  title={Feedback loops with language models drive in-context reward hacking, 2024},
  author={Pan, Alexander and Jones, Erik and Jagadeesan, Meena and Steinhardt, Jacob},
  journal={URL https://arxiv. org/abs/2402.06627},
  year={2024}
}

@article{reasoningefficient,
  title={Training language models to reason efficiently},
  author={Arora, Daman and Zanette, Andrea},
  journal={arXiv preprint arXiv:2502.04463},
  year={2025}
}

@article{thinkprune,
  title={Thinkprune: Pruning long chain-of-thought of llms via reinforcement learning},
  author={Hou, Bairu and Zhang, Yang and Ji, Jiabao and Liu, Yujian and Qian, Kaizhi and Andreas, Jacob and Chang, Shiyu},
  journal={arXiv preprint arXiv:2504.01296},
  year={2025}
}

\clearpage
\newpage

\appendix

\section{Related Work}\label{sec:Related Work}

To train hybrid reasoning models, current methods fall into two categories: SFT-based and RL-only. 

\subsection{SFT-Based Training Method}  

The first method uses SFT to ensure a model possesses the ability to automatically select between thinking and non-thinking modes according to the difficulty of the query \citep{chen2025pangu, jiang2025_LHRMs, lou2025adacot, zhan2025kat}. Specifically, this method fine-tunes the model on a specially constructed dataset in which simple queries are paired with the non-thinking mode while complex queries are paired with the thinking mode. In some cases, RL is applied following SFT to further increase this capability~\citep{jiang2025_LHRMs,lou2025adacot,zhan2025kat}. However, a primary drawback of this method is that it heavily relies on large-scale, high-quality labeled data. The process of curating the SFT dataset---namely classifying queries by difficulty and generating corresponding high-quality responses for both modes---remains highly computationally intensive. In addition, performing SFT with the obtained SFT dataset is likewise highly computationally intensive.

\subsection{RL-Only Training Method}\label{subsec:RL-only training method}

To avoid the substantial computational overhead associated with SFT, other works rely solely on RL to teach the model to automatically select between thinking and non-thinking modes, such as Thinkless, AutoThink, and AdaptThink \citep{fang2025thinkless,tu2025learning,zhang2025adaptthink}. Specifically, the model is incentivized with a higher reward for correct non-thinking answers than for correct thinking answers. Unfortunately, this training method is highly susceptible to reward hacking problem~\citep{skalse2022defining,pan2402feedback}. Precisely, this training method often rely on superficial signals—such as special control tokens—to assign rewards, without visibility into the model's internal computational process. Consequently, the model might learn to alternative the tags to receive higher reward for generating the non-thinking control tokens while still performing thinking internally. 

To mitigate this reward-hacking problem, \citet{fang2025thinkless} augment an SFT process. Unlike the SFT-based training method mentioned previously, this SFT is used solely to fix the model's outputs in thinking and non-thinking modes and does not teach the model to select between these modes based on task difficulty. Although the data-collection process for this SFT is simpler, the computational overhead of this SFT training remains substantial. 

To avoid such high computational overhead, \citet{zhang2025adaptthink} restrict the maximum token usage for the non-thinking mode to ensure the model does not masquerade as non-thinking mode while using the thinking mode to obtain higher returns. Unfortunately, existing work imposes the same maximum token usage for the non-thinking mode across all queries, whereas different queries require different maximum token usage. Specifically, consider a simple query, such as what is $1+1$ equal to, and a complex query, such as an AIME query; even when employing a long CoT to solve the simple query and continuously doing exploration, reflection, and self-verification for the response, token usage may still fail to beat solving the difficult AIME query via a CoT. In this situation, setting the maximum token usage for the non-thinking mode at a low level ensures the identification of reward hacking problem in the non-thinking mode responses to the simple query. However, this method results in almost all non-thinking mode responses to the complex query mode being classified as experiencing reward hacking problem. Conversely, if the maximum token usage is set high enough to distinguish reward hacking problem in the non-thinking mode responses to the complex query effectively, then almost all the non-thinking mode responses to the simple query are classified as not experiencing reward hacking problem. Therefore, applying a uniform maximum token usage for the non-thinking mode across all queries results in poor detection of reward hacking problem, leading only limited mitigation of the reward hacking problem.

Our TNT leverages the solution component of the thinking mode responses to determine the maximum token usage for the non-thinking mode across various queries. Consequently, TNT enhances its ability to identify whether reward hacking problem occurs, thereby effectively mitigating this problem and achieving superior performance.

\section{Addition Experiments}\label{sec:Extended Experiments}

\begin{figure*}[t]
  \centering
   \subfigure{
 \begin{minipage}[b]{0.33\linewidth}
  \includegraphics[width=1\linewidth]{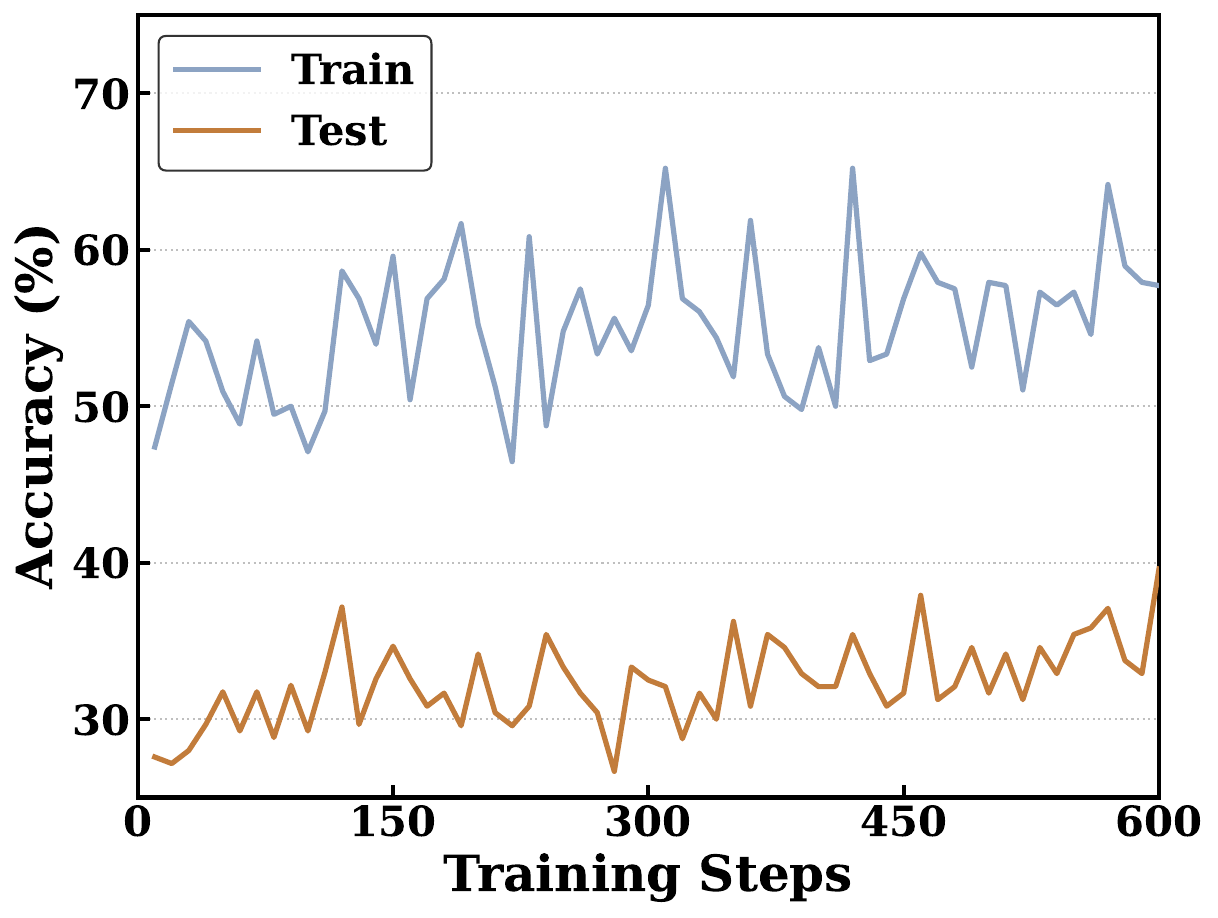}
  \centering 
 \end{minipage}
 \begin{minipage}[b]{0.33\linewidth}
  \includegraphics[width=1\linewidth]{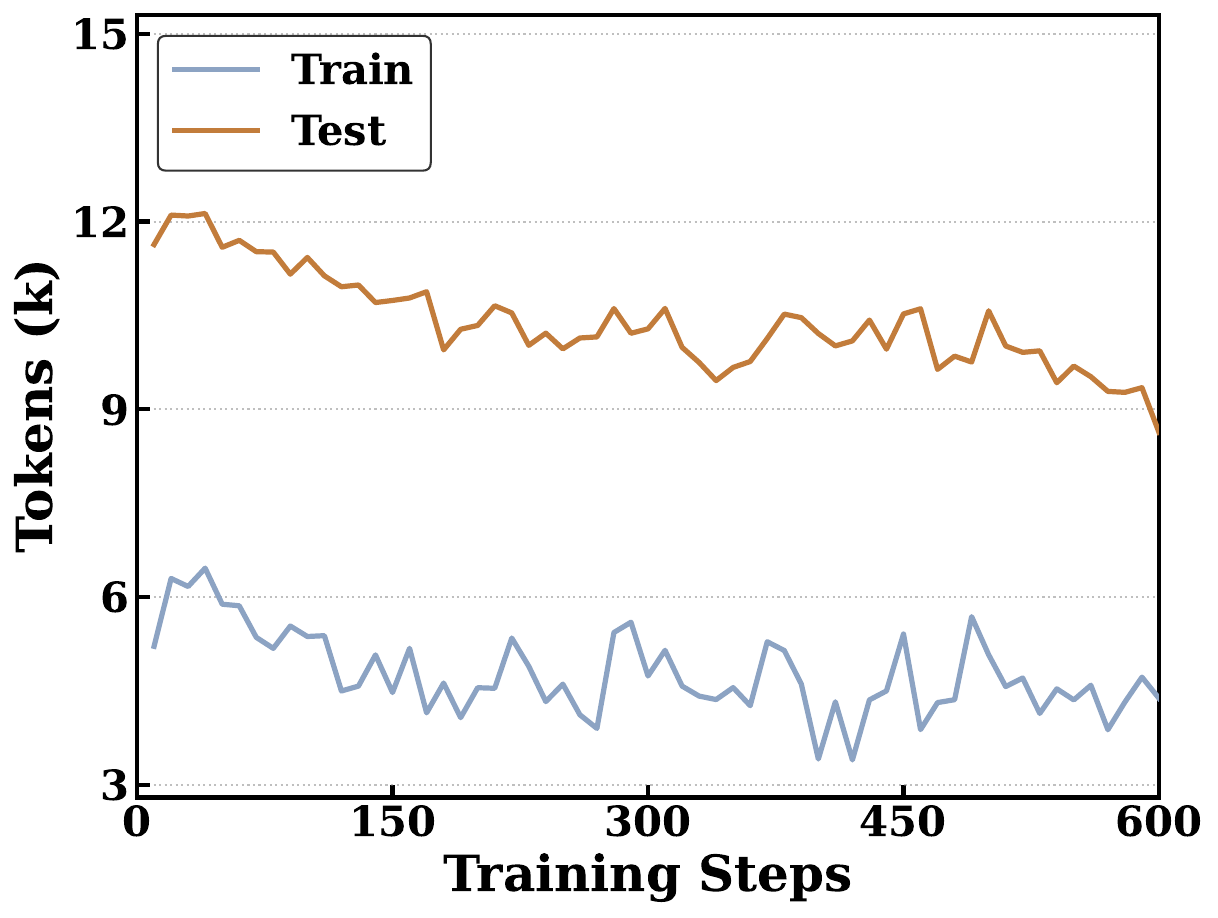}
  \centering  
 \end{minipage}
 \begin{minipage}[b]{0.33\linewidth}
  \includegraphics[width=1\linewidth]{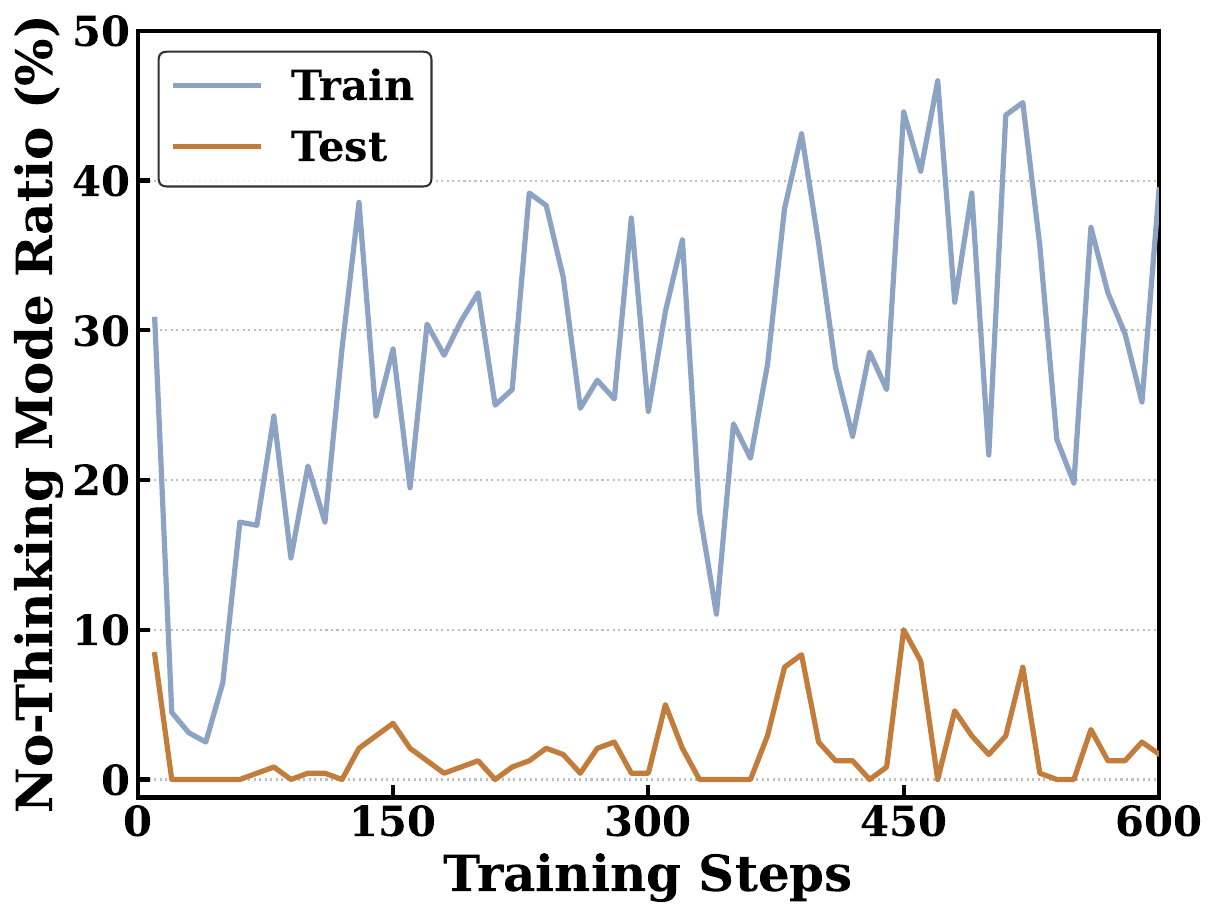}
  \centering 
 \end{minipage}
 }\vspace{-10pt}
 \caption{Accuracy (left), token usage (middle), and non-thinking ratio (right) in RL training on DeepSeek-R1-Distill-Qwen-1.5B. Due to constraints on computational resources, we only use AIME24 as the test dataset.
 }
\label{fig:training dynamics}
  \subfigure{
    \begin{minipage}[b]{0.33\linewidth}
      \includegraphics[width=1\linewidth]{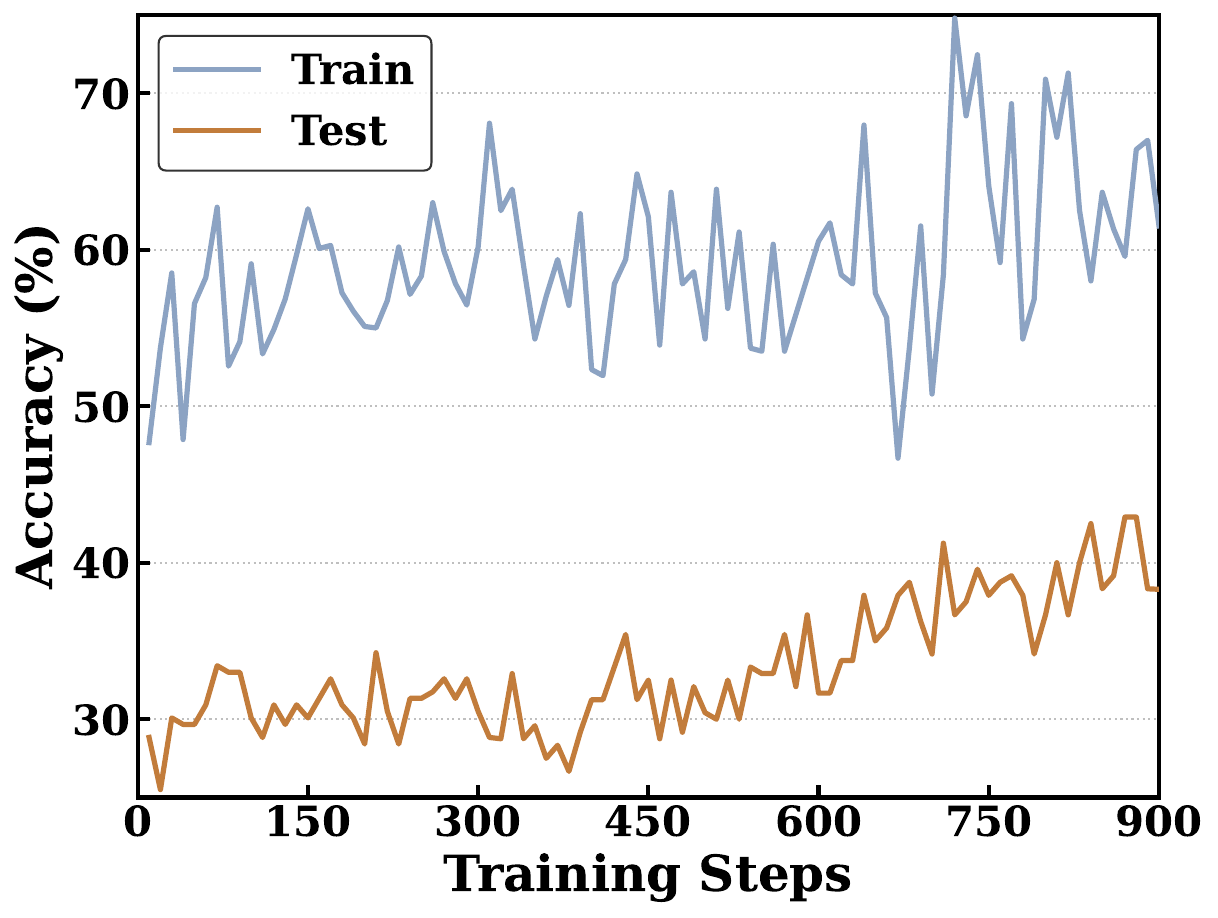}
      \centering
    \end{minipage}
    \begin{minipage}[b]{0.33\linewidth}
      \includegraphics[width=1\linewidth]{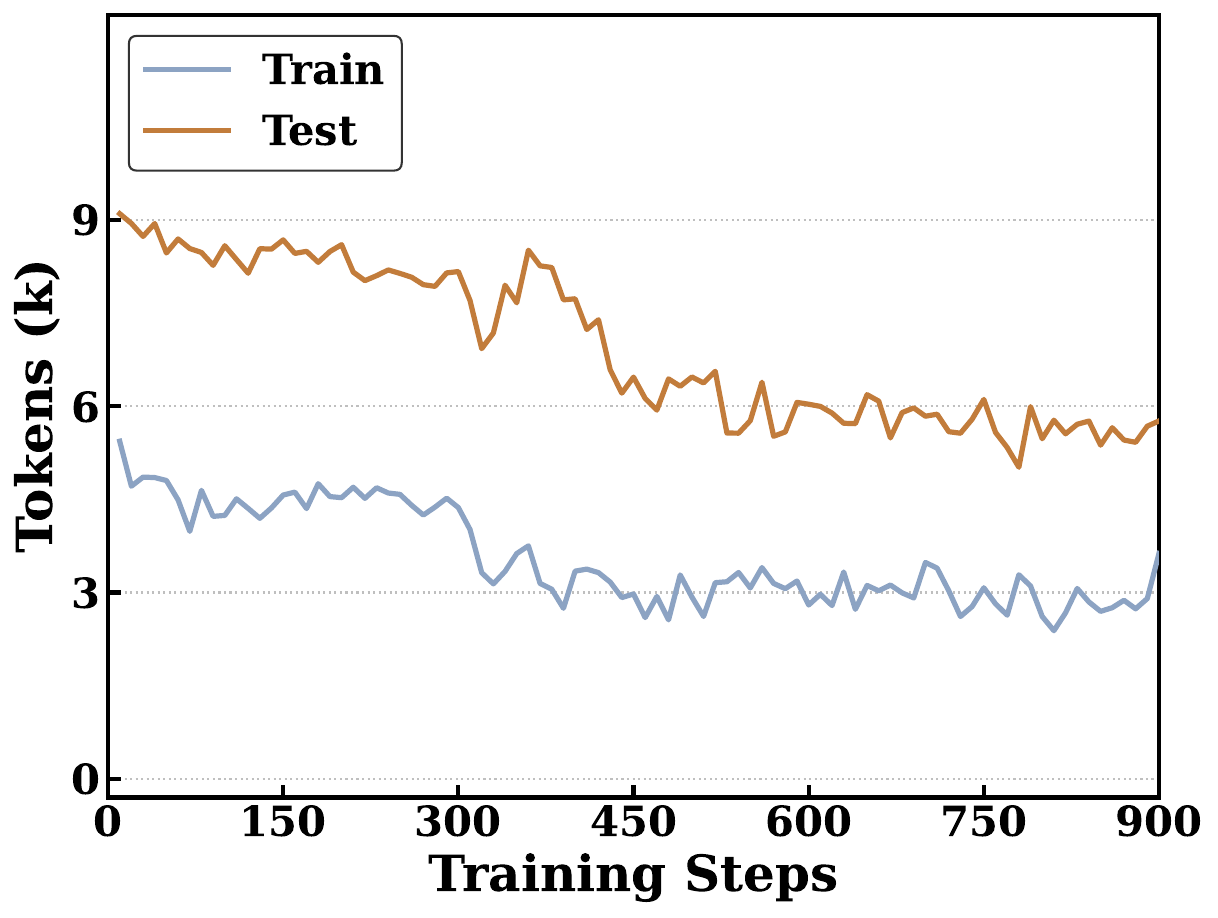}
      \centering
    \end{minipage}
    \begin{minipage}[b]{0.33\linewidth}
      \includegraphics[width=1\linewidth]{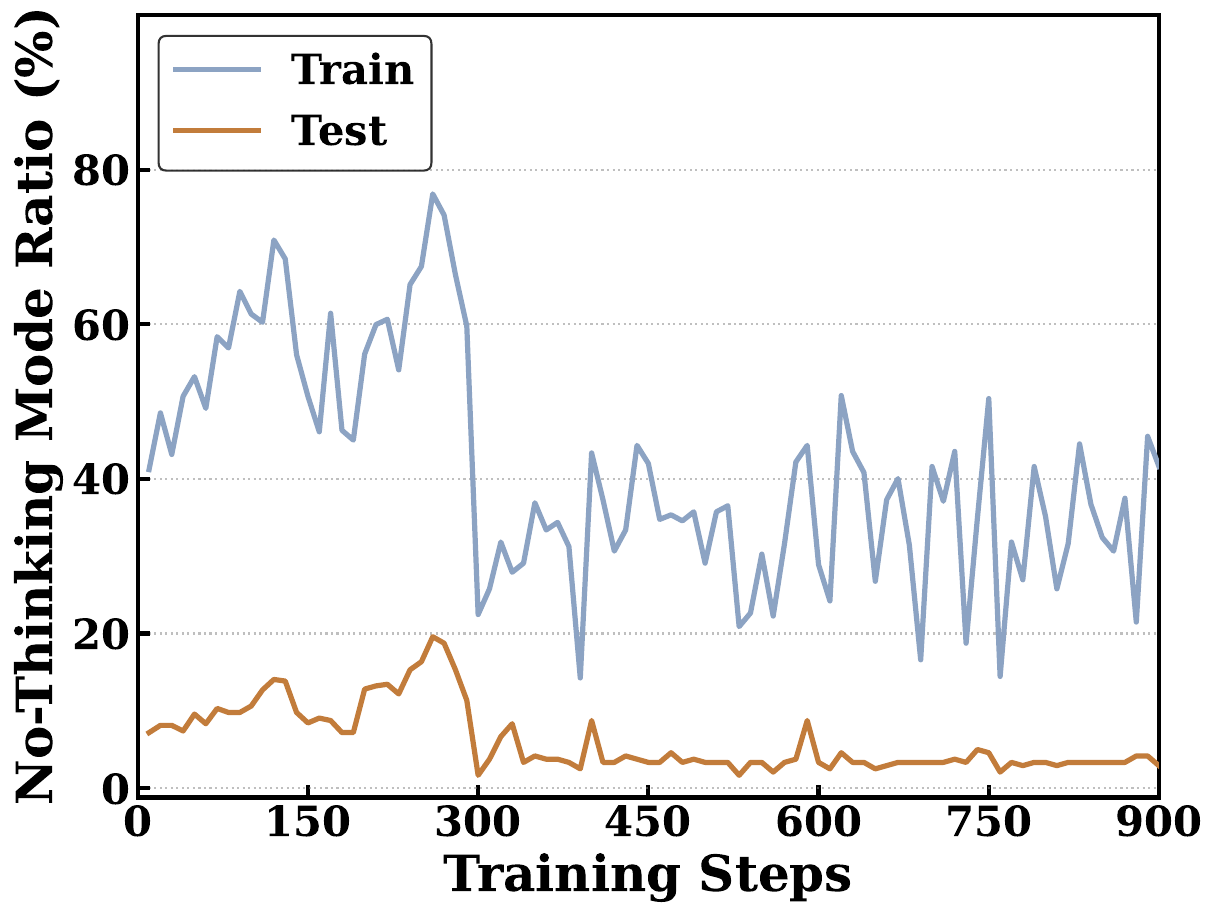}
      \centering
    \end{minipage}
  }
  \vspace{-10pt} 
  \caption{Accuracy (left), token usage (middle), and non-thinking ratio (right) in RL training on DeepScaleR-1.5B. Due to constraints on computational resources, we only use AIME24 as the test dataset.}
  \label{fig:training dynamics deepscaler}
  \vspace{0.2cm} 
  \subfigure{
    \begin{minipage}[b]{0.33\linewidth}
      \includegraphics[width=1\linewidth]{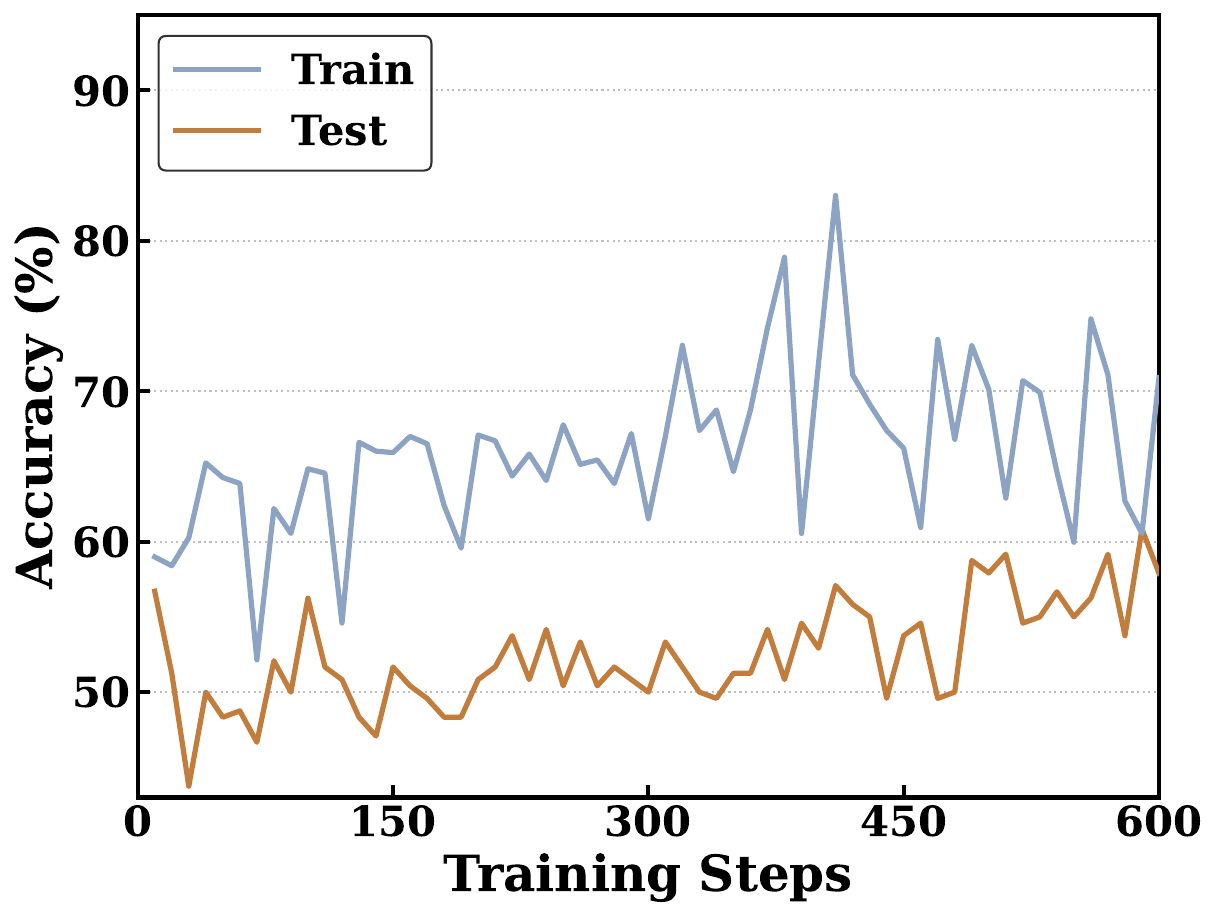}
      \centering
    \end{minipage}
    \begin{minipage}[b]{0.33\linewidth}
      \includegraphics[width=1\linewidth]{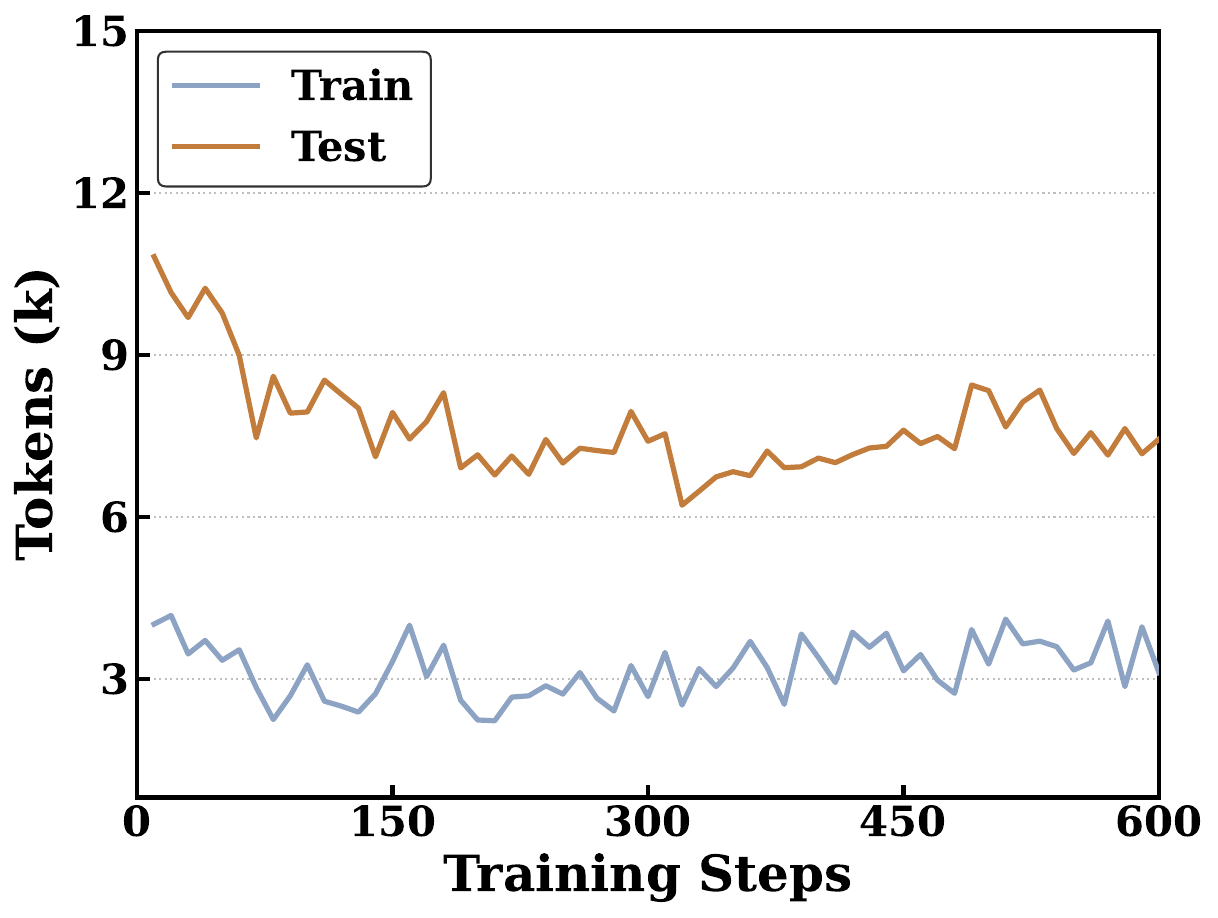}
      \centering
    \end{minipage}
    \begin{minipage}[b]{0.33\linewidth}
      \includegraphics[width=1\linewidth]{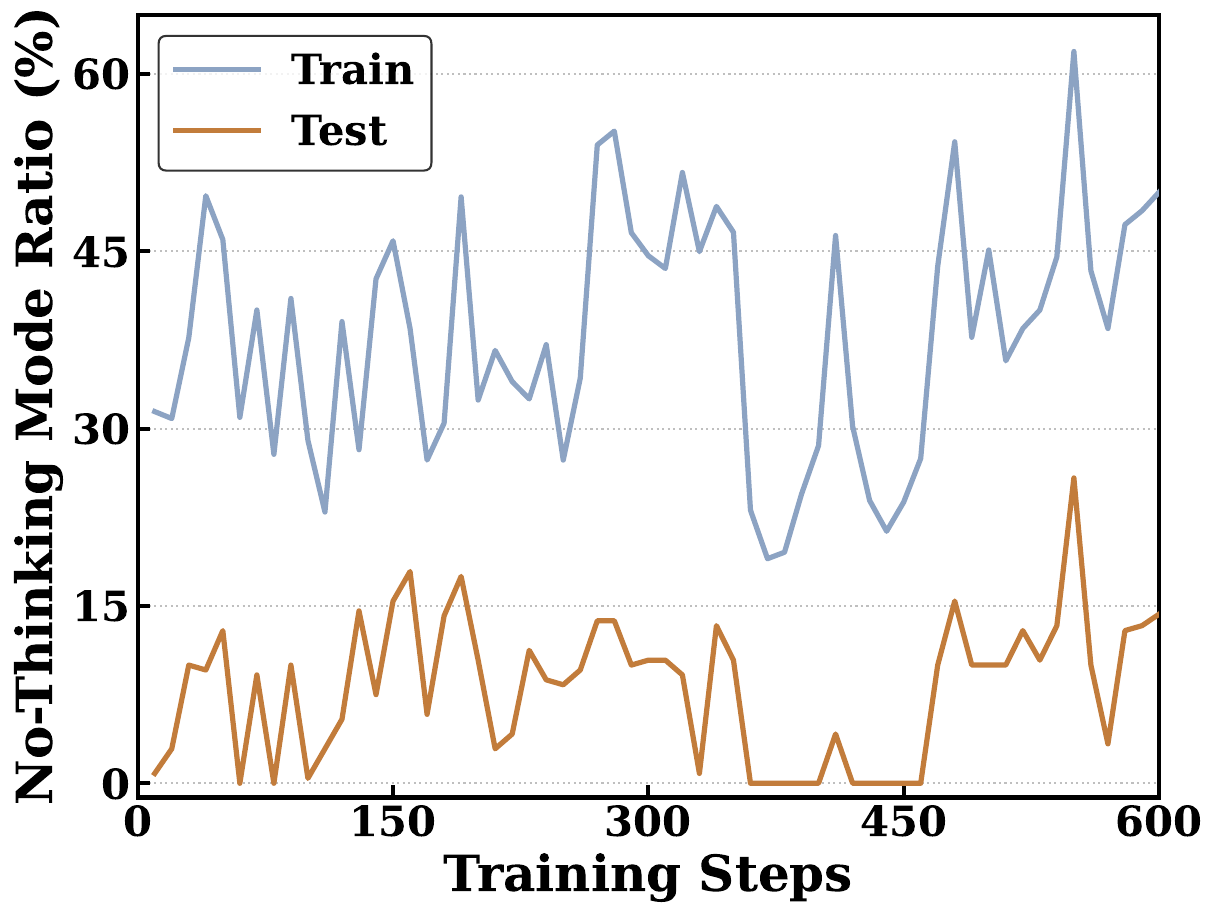}
      \centering
    \end{minipage}
  }
  \vspace{-10pt} 
  \caption{Accuracy (left), token usage (middle), and non-thinking ratio (right) in RL training on DeepSeek-R1-Distill-Qwen-7B. Due to constraints on computational resources, we only use AIME24 as the test dataset.}
  \label{fig:training dynamics 7b}
  
  \vspace{-0.15cm} 
\end{figure*}

\begin{table*}[!t]
\centering
\small
\newcolumntype{Y}{>{\centering\arraybackslash}X}
\setlength{\tabcolsep}{1pt} 
\renewcommand{\arraystretch}{1.2}

\begin{tabularx}{\linewidth}{l l *{13}{Y}} 
\toprule
& \multicolumn{1}{c}{\multirow{2}{*}{\textbf{Models}}} & \multicolumn{6}{c}{\textbf{Accuracy (\%) $\uparrow$}} & \multicolumn{6}{c}{\textbf{Token Usage $\downarrow$}} & \multirow{2}{*}{\textbf{TE}} \\
\cmidrule(lr){3-8} \cmidrule(lr){9-14}
& & \scriptsize AIME24 & \scriptsize AIME25 & \scriptsize Minerva & \scriptsize AMC23 & \scriptsize Olym. & \textbf{\scriptsize AVG} & \scriptsize AIME24 & \scriptsize AIME25 & \scriptsize Minerva & \scriptsize AMC23 & \scriptsize Olym. & \textbf{\scriptsize AVG} & \\
\midrule
\multicolumn{15}{c}{\textit{\textbf{Base Model: DeepScaleR-1.5B}}} \\
\midrule
\rowcolor{gray!8}
& Base Model & \underline{41.9} & \underline{30.6} & 28.4 & 71.4 & 42.9 & 43.0 & 9123 & 9113 & 4175 & 5275 & 5375 & 6612 & 0.53 \\
\rowcolor{cyan!6}
& Auto-S1 & 32.3 & 24.8 & 26.4 & 65.2 & 41.5 & 38.0 & 12819 & 11779 & 5203 & 5467 & 7068 & 8467 & 0.41 \\
\rowcolor{cyan!6}
& Auto-S2 & 39.8 & \textbf{31.0} & \underline{31.6} & \underline{72.2} & \textbf{44.0} & \textbf{43.7} & \underline{8651} & \underline{8772} & 4467 & 5351 & 5434 & 6535 & \underline{0.54} \\
\rowcolor{cyan!6}
& Auto-S3 & \textbf{42.0} & 30.0 & 30.7 & 71.4 & 42.9 & 43.4 & 9030 & 9204 & \underline{3819} & \underline{4682} & \underline{5073} & \underline{6362} & \underline{0.54} \\
\rowcolor{orange!10}
& \textit{\textbf{TNT (Ours)}} & 39.6 & 30.2 & \textbf{31.8} & \textbf{72.7} & \underline{43.1} & \underline{43.5} & \textbf{5479} & \textbf{5216} & \textbf{2402} & \textbf{3208} & \textbf{3236} & \textbf{3908} & \textbf{0.70} \\
\midrule
\multicolumn{15}{c}{\textit{\textbf{Base Model: DeepSeek-R1-Distill-Qwen-7B}}} \\
\midrule
\rowcolor{gray!8} 
& Base Model & \underline{57.6} & 38.4 & \textbf{38.0} & 81.7 & \underline{49.5} & \underline{53.0} & 12804 & 14786 & 5028 & 7733 & 8812 & 9833 & 0.53 \\
\rowcolor{cyan!6}
& Adapt-5 & 56.1 & 37.1 & 34.0 & 81.0 & 48.3 & 51.3 & 9966 & 11553 & 2528 & 5284 & 6381 & 7142 & 0.61 \\
\rowcolor{cyan!6}
& Auto-S1 & 51.7 & 36.7 & 31.8 & 77.7 & 46.9 & 49.0 & 8625 & 9575 & \underline{1763} & \underline{4517} & \underline{5117} & \underline{5919} & 0.64 \\
\rowcolor{cyan!6}
& Auto-S2 & 55.0 & \underline{40.1} & \underline{37.6} & \underline{82.4} & 49.4 & 52.9 & 9078 & 9707 & 3019 & 4710 & 5723 & 6447 & 0.66 \\
\rowcolor{cyan!6}
& Auto-S3 & 53.7 & \underline{40.1} & 37.4 & 81.3 & 49.2 & 52.3 & \underline{8210} & \underline{9446} & 2759 & 4861 & 5549 & 6165 & \underline{0.67} \\
\rowcolor{orange!10} 
& \textit{\textbf{TNT (Ours)}} & \textbf{57.9} & \textbf{41.7} & 36.5 & \textbf{83.9} & \textbf{51.1} & \textbf{54.2} & \textbf{6855} & \textbf{7595} & \textbf{1628} & \textbf{3619} & \textbf{4103} & \textbf{4760} & \textbf{0.79} \\
\bottomrule
\end{tabularx}
\vspace{-0.3cm}
\caption{Comparison of accuracy, token usage, and TE on mathematical benchmarks across hybrid reasoning models with base models DeepScaleR-1.5B and DeepSeek-R1-Distill-Qwen-7B. The best and second results are bolded and underlined, respectively. We did not compare ThinkLess on DeepScaler-1.5B and DeepSeek-R1-Distill-Qwen-7B, nor did we compare AdaptThink on DeepSeek-R1-Distill-Qwen-7B, due to the lack of open-source checkpoints for these methods on the corresponding models. Especially, we exclusively present adapt-5 for DeepSeek-R1-Distill-Qwen-7B, since it serves as the only open-source checkpoint provided by AdaptThink based on this base model.}
\label{tab:main-7B}
\end{table*}

\begin{table*}[!t]
\centering
\small
\newcolumntype{Y}{>{\centering\arraybackslash}X}
\setlength{\tabcolsep}{1pt} 
\renewcommand{\arraystretch}{1.2}
\begin{tabularx}{\linewidth}{l l *{12}{Y}} 
\toprule
& \multicolumn{1}{c}{\multirow{2}{*}{\textbf{Models}}} & \multicolumn{6}{c}{\textbf{Non-Thinking Mode Tokens}} & \multicolumn{6}{c}{\textbf{Thinking Mode Tokens}} \\
\cmidrule(lr){3-8} \cmidrule(lr){9-14} 
& & \scriptsize AIME24 & \scriptsize AIME25 & \scriptsize Minerva & \scriptsize AMC23 & \scriptsize Olym. & \textbf{\scriptsize AVG} & \scriptsize AIME24 & \scriptsize AIME25 & \scriptsize Minerva & \scriptsize AMC23 & \scriptsize Olym. & \textbf{\scriptsize AVG} \\
\midrule
\rowcolor{orange!10} 
& TNT-DeepSeek-1.5B & 995 & 795 & 601 & 859 & 937 & 837 & 8633 & 8325 & 3475 & 7159 & 6086 & 6736 \\
\rowcolor{orange!10} 
& TNT-DeepScaleR-1.5B & 2779 & - & 812 & 967 & 1206 & 1153 & 5663 & 5216 & 2475 & 4020 & 3775 & 4230 \\
\rowcolor{orange!10} 
& TNT-DeepSeek-7B & 1552 & 1359 & 702 & 1214 & 1185 & 1202 & 7738 & 8435 & 2924 & 6069 & 6327 & 6299 \\
\bottomrule
\end{tabularx}
\vspace{-0.3cm}
\caption{No-thinking and thinking mode token usage of TNT on mathematical benchmarks. TNT-DeepSeek-1.5B/7B refer to TNT with base mode DeepSeek-R1-Distill-Qwen-1.5B/7B and TNT-DeepScaleR-1.5B refer to TNT with base mode DeepScaleR-1.5B. We also follow these abbreviation conventions in the table below.}
\vspace{-0.4cm}
\label{tab:three model token_usage}
\end{table*}

\begin{table}[t]
\centering
\small
\newcolumntype{Y}{>{\centering\arraybackslash}X}
\setlength{\tabcolsep}{1pt} 
\renewcommand{\arraystretch}{1.2} 

\begin{tabularx}{\linewidth}{l *{5}{Y}} 
\toprule
\multicolumn{1}{c}{\multirow{2}{*}{\textbf{Models}}} & \multicolumn{5}{c}{\textbf{No-Thinking Mode Ratio (\%)}} \\
\cmidrule(lr){2-6}
& \scriptsize AIME24 & \scriptsize AIME25 & \scriptsize Minerva & \scriptsize AMC23 & \scriptsize Olym. \\
\midrule
\rowcolor{orange!10} 
TNT-DeepSeek-1.5B & 1.7 & 0.4 & 11.3 & 30.7 & 20.7 \\
\rowcolor{orange!10} 
TNT-DeepScaleR-1.5B & 2.9 & 0 & 4.4 & 26.6 & 21.0 \\
\rowcolor{orange!10} 
TNT-DeepSeek-7B & 14.3 & 11.9 & 58.3 & 50.5 & 43.2 \\
\bottomrule
\end{tabularx}
\vspace{-0.2cm}
\caption{Non-thinking mode ratio of TNT on mathematical benchmarks across base models.}
\vspace{-0.5cm}
\label{tab:No-Thinking Mode TNT}
\end{table}

\begin{table*}[!t]
\centering
\small
\newcolumntype{Y}{>{\centering\arraybackslash}X}
\setlength{\tabcolsep}{1pt}
\renewcommand{\arraystretch}{1.2}

\begin{tabularx}{\linewidth}{l l *{13}{Y}}
\toprule
& \multicolumn{1}{c}{\multirow{2}{*}{\textbf{Models}}} & \multicolumn{6}{c}{\textbf{Accuracy (\%) $\uparrow$}} & \multicolumn{6}{c}{\textbf{Token Usage $\downarrow$}} & \multirow{2}{*}{\textbf{TE}} \\
\cmidrule(lr){3-8} \cmidrule(lr){9-14}
& & \scriptsize AIME24 & \scriptsize AIME25 & \scriptsize Minerva & \scriptsize AMC23 & \scriptsize Olym. & \textbf{\scriptsize AVG} & \scriptsize AIME24 & \scriptsize AIME25 & \scriptsize Minerva & \scriptsize AMC23 & \scriptsize Olym. & \textbf{\scriptsize AVG} & \\
\midrule

\multicolumn{15}{c}{\textit{\textbf{Base Model: DeepSeek-R1-Distill-Qwen-1.5B}}} \\
\midrule
\rowcolor{gray!8}
& Base Model & 28.6 & 24.6 & 26.2 & 62.1 & \underline{43.6} & 37.0 & 16865 & 16464 & 7490 & 11050 & 11808 & 12736 & 0.33 \\
\rowcolor{green!6}
& RE-0 & \underline{31.9} & \underline{24.8} & \underline{28.3} & \underline{65.6} & 40.6 & \underline{38.2} & 13404 & 12807 & 4666 & 7912 & 8493 & 9456 & 0.39 \\
\rowcolor{green!6}
& RE-1 & 30.3 & 24.0 & 25.8 & 64.3 & 38.2 & 36.5 & 12660 & 12474 & 3633 & 6827 & 7665 & 8652 & 0.39 \\
\rowcolor{green!6}
& RE-4 & 28.9 & 21.4 & 19.7 & 62.4 & 33.5 & 33.2 & 9467 & 8960 & \textbf{1444} & 4902 & 5424 & 6039 & 0.43 \\
\rowcolor{green!6}
& TP-4k & 29.2 & 21.2 & \underline{28.3} & 65.3 & 39.4 & 36.7 & 8154 & 8261 & 3353 & 4843 & 5586 & 6039 & 0.47 \\
\rowcolor{green!6}
& TP-2k & 27.3 & 22.2 & 26.7 & 64.4 & 38.2 & 35.8 & \underline{7541} & \underline{7140} & \underline{2430} & \underline{4462} & \underline{4815} & \underline{5278} & 0.49 \\
\rowcolor{green!6}
& TP-iter2k & 27.3 & 20.6 & 26.2 & 65.1 & 39.0 & 35.6 & \textbf{6990} & \textbf{5947} & 2805 & \textbf{3602} & \textbf{3780} & \textbf{4625} & \underline{0.52} \\
\rowcolor{orange!10}
& \textit{\textbf{TNT (Ours)}} & \textbf{37.7} & \textbf{26.1} & \textbf{28.7} & \textbf{65.9} & \textbf{46.4} & \textbf{41.0} & 8537 & 8252 & 3150 & 4508 & 5020 & 5893 & \textbf{0.53} \\
\midrule

\multicolumn{15}{c}{\textit{\textbf{Base Model: DeepScaleR-1.5B}}} \\
\midrule
\rowcolor{gray!8}
& Base Model & \textbf{41.9} & \textbf{30.6} & 28.4 & 71.4 & \underline{42.9} & \underline{43.0} & 9123 & 9113 & 4175 & 5275 & 5375 & 6612 & 0.53 \\
\rowcolor{green!6}
& TP-4k & 37.9 & 28.8 & \textbf{32.3} & \underline{72.6} & 42.3 & 42.8 & 6592 & 6095 & 3181 & 3907 & 4070 & 4769 & 0.62 \\
\rowcolor{green!6}
& TP-2k & 35.8 & 26.9 & 29.8 & 69.9 & \underline{42.9} & 41.1 & 5687 & 5315 & \underline{2482} & 3458 & 3575 & 4103 & 0.64 \\
\rowcolor{green!6}
& TP-iter2k & 36.6 & 28.5 & 31.3 & 71.4 & 42.3 & 42.0 & \underline{5495} & \underline{5074} & 2595 & \underline{3355} & \underline{3498} & \underline{4003} & \underline{0.66} \\
\rowcolor{orange!10}
& \textit{\textbf{TNT (Ours)}} & \underline{39.6} & \underline{30.2} & \underline{31.8} & \textbf{72.7} & \textbf{43.1} & \textbf{43.5} & \textbf{5479} & \textbf{5216} & \textbf{2402} & \textbf{3208} & \textbf{3236} & \textbf{3908} & \textbf{0.70} \\
\midrule

\multicolumn{15}{c}{\textit{\textbf{Base Model: DeepSeek-R1-Distill-Qwen-7B}}} \\
\midrule
\rowcolor{gray!8}
& Base Model & 57.6 & 38.4 & \textbf{38.0} & 81.7 & 49.5 & 53.0 & 12804 & 14786 & 5028 & 7733 & 8812 & 9833 & 0.53 \\
\rowcolor{green!6}
& RE-0 & \textbf{59.5} & \underline{40.6} & \underline{37.5} & \underline{82.4} & \underline{49.7} & \underline{53.9} & 11171 & 13217 & 4074 & 6703 & 7853 & 8604 & 0.58 \\
\rowcolor{green!6}
& RE-1 & 52.9 & 40.5 & 37.2 & 81.3 & 48.9 & 52.2 & 10696 & 11420 & 2972 & 5827 & 6761 & 7535 & 0.60 \\
\rowcolor{green!6}
& RE-4 & 55.2 & 38.8 & 32.9 & 81.0 & 48.3 & 51.2 & \underline{9775} & \underline{10988} & \underline{2611} & \underline{5086} & \underline{6088} & \underline{6910} & \underline{0.62} \\
\rowcolor{orange!10}
& \textit{\textbf{TNT (Ours)}} & \underline{57.9} & \textbf{41.7} & 36.5 & \textbf{83.9} & \textbf{51.1} & \textbf{54.2} & \textbf{6855} & \textbf{7595} & \textbf{1628} & \textbf{3619} & \textbf{4103} & \textbf{4760} & \textbf{0.79} \\
\bottomrule

\end{tabularx}
\vspace{5pt}
\caption{Comparison of accuracy, token usage, and TE on mathematical benchmarks between TNT and CoT compression models with base models DeepSeek-R1-Distill-Qwen-1.5B, DeepScaleR-1.5B, and DeepSeek-R1-Distill-Qwen-7B. \textcolor{gray}{Gray} represents the base models, \textcolor{green}{green} denotes efficient reasoning models, and \textcolor{orange}{orange} represents TNT. RE-$x$ refers to Efficient-Reasoning with $\alpha=x*0.01$. TP-$x$ refers to ThinkPrune-$x$. The best and second results are bolded and underlined, respectively. We did not compare RE on DeepScaler-1.5B, nor did we compare TP on DeepSeek-R1-Distill-Qwen-7B, due to the lack of open-source checkpoints for these methods on the corresponding models.}
\label{tab:main-compression}
\end{table*}

\subsection{Experiment on DeepScaleR-1.5B and DeepSeek-R1-Distill-Qwen-7B}\label{subsec:deepscaler and 7B}

To further verify the universality and scalability of TNT, we extended our experiments to DeepScaleR-1.5B and DeepSeek-R1-Distill-Qwen-7B. As shown in Table \ref{tab:main-7B}, TNT demonstrates consistent superiority in TE across both base models, significantly reducing computational costs while maintaining competitive performance. Specifically, TNT reduces average token usage by approximately 41\% for DeepScaleR-1.5B and over 51\% for DeepSeek-R1-Distill-Qwen-7B compared to their respective base models. Importantly, this substantial reduction in token usage does not compromise model performance; TNT consistently matches or surpasses the accuracy of the base models and robust baselines. The training dynamics illustrated in Figure \ref{fig:training dynamics deepscaler} and Figure \ref{fig:training dynamics 7b} align perfectly with our observations on the 1.5B model, where accuracy steadily improves on the test set while token consumption naturally decreases throughout the RL process. This confirms that the efficacy of our TNT is robust across different model scales and initial distributions without requiring explicit length penalties.

While the overarching trend of enhanced efficiency is consistent, the performance gains manifest with slight nuances between the two models. In the case of DeepScaleR-1.5B, which is already highly optimized for mathematical reasoning, the primary contribution of TNT lies in aggressive efficiency optimization. Although its average accuracy of 43.5\% is comparable to the 43.7\% achieved by AutoThink-Stage2, TNT achieves this result using only 3908 average tokens compared to the 6535 tokens required by the latter. This results in a superior TE of 0.70 compared to the 0.54 score for AutoThink-Stage2 and AutoThink-Stage3. And for DeepSeek-R1-Distill-Qwen-7B, TNT achieves a comprehensive improvement by securing both the highest average accuracy of 54.2\%, which surpasses the base model by 1.2\%, and the lowest token usage among all compared methods. Consequently, it attains a remarkable TE score of 0.79, significantly outperforming the best AutoThink-Stage3 baseline score of 0.67, demonstrating that TNT outperforms baselines on strong reasoners, with the advantage becoming more pronounced as model capability increases.

We analyze the token usage across different base models to investigate the impact of model capacity on reasoning length. 
As shown in Table~\ref{tab:three model token_usage}, the token consumption in the non-thinking mode is not statically bounded; instead, it exhibits significant behavioral differences depending on the base model's strength. We analyze the impact of base model scale and initialization on the mode selection mechanism as shown in Table \ref{tab:No-Thinking Mode TNT}. A clear positive correlation is observed between model parameter size and the utilization of the no-thinking mode. When scaling from 1.5B to 7B parameters, the TNT framework exhibits a significantly higher propensity to bypass the reasoning process; for instance, the no-thinking ratio on the Minerva benchmark surges from 11.3\% with TNT-DeepSeek-1.5B to 58.3\% with TNT-DeepSeek-7B. This suggests that larger models, possessing richer internalized knowledge, can more frequently solve problems directly without invoking the extensive reasoning chain. Conversely, the performance of TNT-DeepScaleR-1.5B closely mirrors that of TNT-DeepSeek-1.5B across all datasets. Although DeepScaleR is a derivative of DeepSeek-1.5B optimized for mathematical reasoning, its training process does not explicitly target or enhance direct answering (no-thinking) capabilities. Consequently, the intrinsic non-reasoning proficiency remains unchanged, resulting in a static distribution of mode selection ratios comparable to the original base model.

\subsection{Comparison with CoT Compression Methods}\label{subsec:cot compression}

To assess the competitiveness of TNT against recent CoT compression strategies, we compare it with two prominent methods: Efficient-Reasoning (RE)~\citep{reasoningefficient}. and ThinkPrune (TP)~\citep{thinkprune}.. As detailed in Table \ref{tab:main-compression}, TNT consistently outperforms both methods in terms of TE across all three base models, demonstrating that our hybrid reasoning model training method offers a superior trade-off between accuracy and computational cost. Unlike standard CoT compression methods that often degrade performance in exchange for reduced token usage, TNT frequently enhances accuracy while simultaneously achieving deeper reductions in token usage. For instance, while RE and TP models typically show a drop in accuracy as CoT compression intensifies, TNT maintains or even improves upon the base model's performance, achieving the highest TE scores. This establishes TNT not merely as a compression tool, but as a reasoning enhancement framework that fundamentally reorganizes how models allocate computational resources.

Analyzing the specific behaviors on different base models reveals distinct advantages of TNT. On the DeepSeek-R1-Distill-Qwen-1.5B model, TNT achieves a remarkable average accuracy of 41.0\%, which is substantially higher than the 36.7\% achieved by the best performing TP-4k and the 38.2\% of RE-0. While TP-iter2k manages to lower token usage slightly more than TNT, its accuracy suffers significantly, dropping to 35.6\%. In contrast, TNT balances a moderate token usage with superior accuracy, resulting in a leading TE of 0.53. Similarly, on the DeepScaleR-1.5B base, TNT surpasses all TP baselines in both efficiency and effectiveness. It achieves an average token usage of 3908, which is lower than the 4003 tokens used by TP-iter2k, while maintaining a higher average accuracy of 43.5\% compared to 42.0\% for the latter. The advantage is even more pronounced on the larger DeepSeek-R1-Distill-Qwen-7B model, where TNT attains an exceptional TE of 0.79. Here, TNT drastically cuts average token usage to 4760 compared to 6910 for the most aggressive baseline RE-4, all while securing the highest average accuracy of 54.2\%. This confirms that TNT is uniquely capable of scaling efficiency gains without the accuracy penalties commonly associated with traditional length-penalty or pruning methods.

\begin{table}[!t]

\centering
\small
\newcolumntype{Y}{>{\centering\arraybackslash}X}
\setlength{\tabcolsep}{4pt}
\renewcommand{\arraystretch}{1.2}

\begin{tabularx}{\linewidth}{l *{3}{Y}} 
\toprule
\multicolumn{1}{c}{\textbf{Models}} & \textbf{Accuracy (\%)} & \textbf{Token Usage} & \textbf{TE} \\
\midrule

\rowcolor{gray!8}
Base Model & \underline{36.9} & 17343 & 0.28 \\

\rowcolor{cyan!6}
Thinkless & 28.8 & \textbf{5958} & \underline{0.37} \\

\rowcolor{cyan!6}
Adapt-0 & 33.3 & 19388 & 0.23 \\

\rowcolor{cyan!6}
Adapt-2 & 35.9 & 18463 & 0.26 \\

\rowcolor{cyan!6}
Adapt-5 & \textbf{37.4} & 17627 & 0.28 \\

\rowcolor{cyan!6}
Auto-S1 & 33.3 & 9794 & 0.33 \\

\rowcolor{cyan!6}
Auto-S2 & 34.3 & 12374 & 0.30 \\

\rowcolor{cyan!6}
Auto-S3 & 31.3 & 8816 & 0.33 \\

\rowcolor{orange!10} 
\textit{\textbf{TNT (Ours)}} & 35.9 & \underline{8803} & \textbf{0.38} \\
\bottomrule

\end{tabularx}
\vspace{2pt}
\caption{Accuracy, token usage and TE comparison across models on GPQA Diamond benchmark.}
\label{tab:gpqadiamond_metrics}
\centering
\resizebox{0.9\linewidth}{!}{
    \renewcommand{\arraystretch}{1.0} 
    \setlength{\tabcolsep}{5pt}
    \small
    \begin{tabular}{cccc}
    \toprule
         &  \textbf{Step 100} & \textbf{Step 300} & \textbf{Step 500} \\
    \midrule
        RF in Eq. (6) & 3.9 &  17.9 & 94.2  \\
    \midrule
        TNT &  0 &  7.2  & 5.1 \\
    \bottomrule
    \end{tabular}
    }
    \vspace{-5pt}
    \caption{Average probability of thinking-related verbs appearing in non-thinking mode responses across Minerva, AMC23, and Olympiad when using the reward function in \Cref{eq:autothink reward function}. RF refers to reward function.}
    \label{tab:Average probability of thinking-related}
    \vspace{-0.35cm}
\end{table}

\begin{figure}[t]
  \centering
  \includegraphics[width=\linewidth]{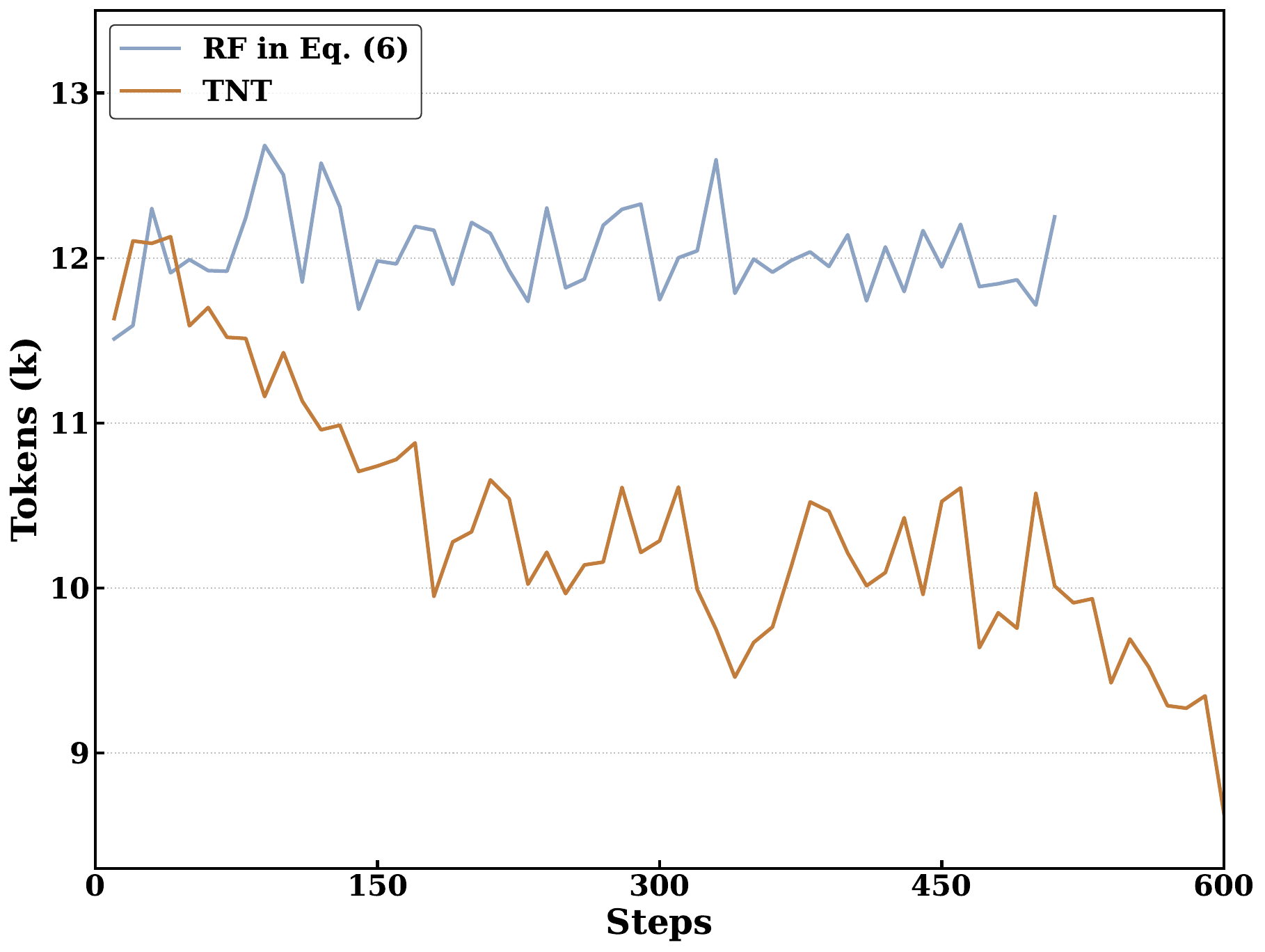}
  \vspace{-13pt}
  \caption{Token usage varies on AIME24 when using the reward function in \Cref{eq:autothink reward function}.}
  \label{fig:Ablation1}
  \centering
  \includegraphics[width=\linewidth]{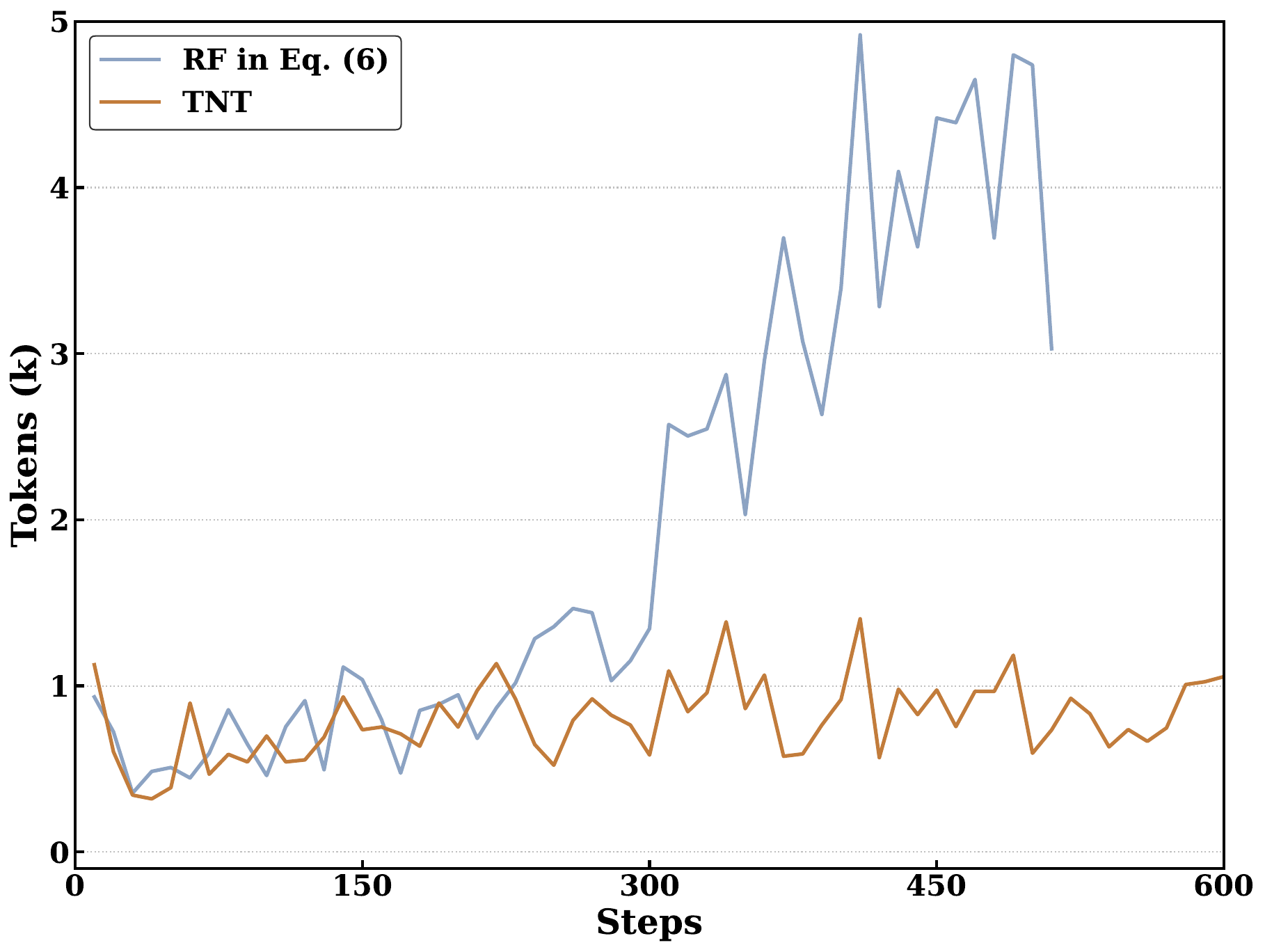}
  \vspace{-13pt}
  \caption{Token usage of the non-thinking mode responses varies on the training dataset when using the reward function in \Cref{eq:autothink reward function}.}
  \label{fig:Ablation2}
  \vspace{-0.35cm}
\end{figure}

\subsection{Experiments in OOD}\label{subsec:Experiments in OOD}

To assess the out-of-distribution (OOD) generalization capabilities of TNT, we conducted experiments on the GPQA Diamond, which features challenging graduate-level queries in biology, physics, and chemistry. As shown in \Cref{tab:gpqadiamond_metrics}, while TNT does not achieve the single best score in accuracy or token usage alone, it strikes a superior balance between performance and computational cost, making it the best overall model. Specifically, TNT achieves the highest TE of 0.38, surpassing all other models. This indicates that it provides high accuracy while remaining highly efficient, a crucial advantage over models that are either slightly more accurate but far more resource-intensive or less accurate. This exceptional balance underscores TNT's robust generalization and practical effectiveness on complex, scientific OOD tasks.

\subsection{Ablation Study}\label{subsec:Ablation Study}

Finally, we conduct tests under the scenario where the components of our reward function related to reward hacking problem are removed. Specifically, the reward function reduces to the naive reward function in \citet{tu2025learning}, as shown in the following:
\begin{equation}\label{eq:autothink reward function}
\begin{aligned}
    & R(x, y_x^k, y_x^*, p(y_x^k),\mathcal{L}_{x}^{\text{N}}) \\ & =
    \begin{cases}
        1, & \text{if } p(y_x^k)=1 \land r(y_x^k, y_x^*)=1, \\
        0, & \text{if } p(y_x^k)=1 \land r(y_x^k, y_x^*)=0, \\
        2, & \text{if } p(y_x^k)=0 \land r(y_x^k, y_x^*)=1,  \\
        -1, & \text{if } p(y_x^k)=0 \land r(y_x^k, y_x^*)=0.  \\
    \end{cases}
\end{aligned}
\end{equation}

We evaluate three key metrics: (i) average probability of thinking-related verbs appearing in responses classified as the non-thinking mode on the testing dataset, (ii) the average token usage across all responses on the testing dataset, and (iii) the average token usage for responses categorized under the non-thinking mode on the training dataset. The average probability of thinking-related verbs appearing in non-thinking mode responses on the testing dataset are presented in \Cref{tab:Average probability of thinking-related}. Due to the high variance in calculating such probabilities for AIME24 and AIME25—originating from the limited sampling of non-thinking mode responses in these datasets—we focus solely on the average probabilities across Minerva, AMC23, and Olympiad. In addition, due to the high computational cost of evaluations on these three datasets, we are able to assess only a limited number of checkpoints. This constraint prevents us from performing evaluations every 10 steps, as shown in the subsequent experimental results. The average token usage across all responses on the testing dataset are presented in \Cref{fig:Ablation1}. Note that the evaluation is conducted solely on AIME24 due to computational resource constraints. Specifically, we cannot evaluate checkpoints on all five mathematical benchmarks every 10 steps. The token usage for responses classified as the non-thinking mode on the training dataset are shown in \Cref{fig:Ablation2}.

We observe that as training progresses, utilizing the reward function shown in \Cref{eq:autothink reward function}, token usage for responses classified as non-thinking mode on the training dataset significantly increases after a certain step. This phenomenon suggests a high probability of reward hacking problem since such token usage is too large. This finding aligns with results presented in \Cref{tab:Average probability of thinking-related}, where TNT’s responses seldom exhibit reward hacking problem, whereas models using the reward function in \Cref{eq:autothink reward function} frequently demonstrate such behavior. Consequently, the elevated probability of reward hacking problem leads to a persistent lack of reduction in token usage for responses on AIME24.

\section{A Case Study}\label{sec:A Case Study}

\begin{figure*}[t]
  \centering
  \includegraphics[width=1.0\textwidth]{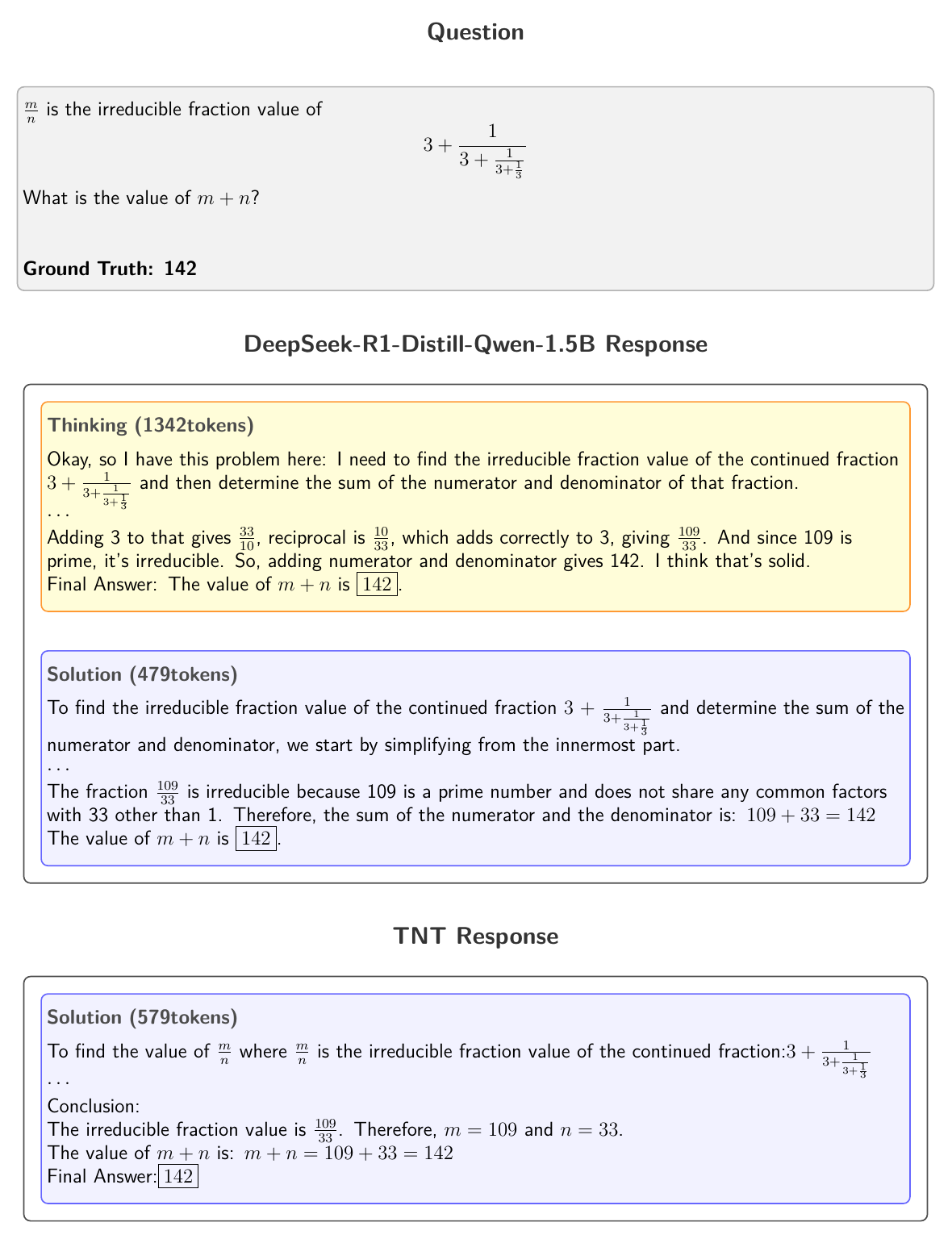}
  \caption{For a easy query, TNT can solve it by switching to the non-thinking mode to use fewer tokens compared to DeepSeek-R1-Distill-Qwen-1.5B.}
  \label{fig:Easy-Case}
  \vspace{-0.15cm}
\end{figure*}

\begin{figure*}[t]
  \centering
  \includegraphics[width=\textwidth]{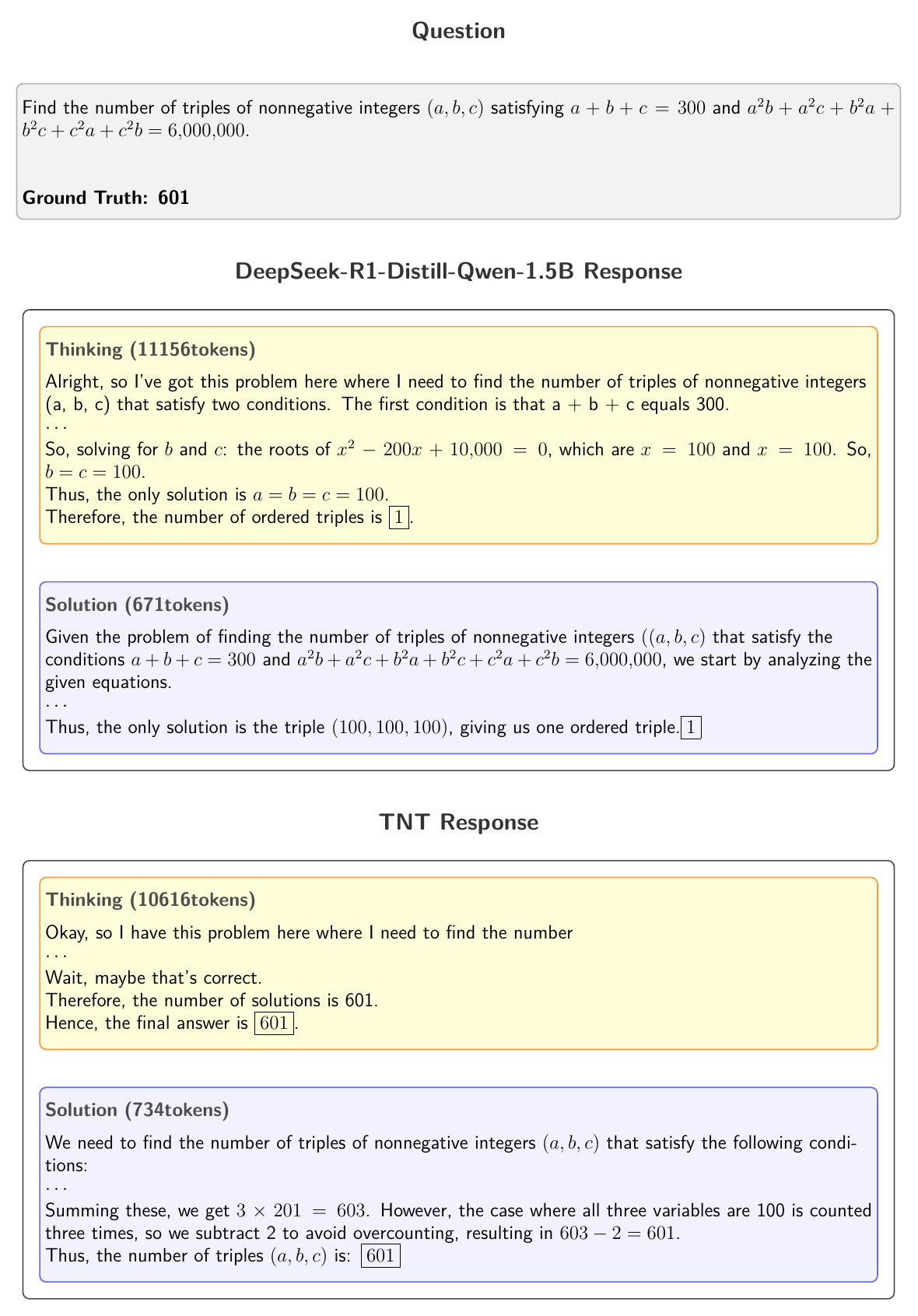}
  \caption{For a hard query, TNT is able to solve it correctly through careful thinking, while DeepSeek-R1-Distill-Qwen-1.5B was not.}
  \label{fig:Hard-Case}
  \vspace{-0.15cm}
\end{figure*}

To provide a more intuitive and qualitative understanding of TNT, we present a series of case studies comparing TNT with the  DeepSeek-R1-Distill-Qwen-1.5B baseline across mathematical queries of varying difficulty. These cases, illustrated in Figures \ref{fig:Easy-Case} and \ref{fig:Hard-Case}, demonstrate TNT's ability to dynamically engage in the thinking or non-thinking mode. Our analysis highlights that for the simple query, TNT opts for the non-thinking mode that is resource-efficient; for the complex query that cause the baseline to fail, TNT engages in the thinking mode to arrive at the correct solution. This adaptive behavior underscores TNT's superior flexibility and effectiveness in solving a wide spectrum of queries.

\end{document}